\RequirePackage{fix-cm}
\documentclass[]{interact}

\usepackage{epstopdf}
\usepackage{float}
\usepackage{booktabs,xltabular,array,ragged2e}
\usepackage[table]{xcolor}
\usepackage[T1]{fontenc}

\usepackage[natbibapa,nodoi]{apacite}
\theoremstyle{plain}

\theoremstyle{definition}

\theoremstyle{remark}

\begin{document}

\articletype{REVIEW}

\title{Active Learning for Biodiversity Monitoring: From Label Efficiency to Reliable Ecological Inference}

\author{
    \name{Ben McEwen\textsuperscript{a}\thanks{CONTACT Ben McEwen. Email: b.j.mcewen@uva.nl}, Shiqi Zhang\textsuperscript{b} and Dan Stowell\textsuperscript{c,d}}
    \affil{\textsuperscript{a}Institute for Biodiversity and Ecosystem Dynamics (IBED), University of Amsterdam, Amsterdam, the Netherlands}
    \affil{\textsuperscript{b}Faculty of Information Technology and Communication Sciences, Tampere University, Finland}
    \affil{\textsuperscript{c}Leiden Institute of Advanced Computer Science, Leiden University, The Netherlands}
    \affil{\textsuperscript{d}Naturalis Biodiversity Centre, Leiden, The Netherlands}
}

\maketitle

\begin{abstract}

    Limited expert annotation capacity is a pervasive constraint in biodiversity monitoring. Passive acoustic recorders and camera traps generate raw data faster than experts can analyse them. Machine learning (ML) models can process these data at scale, but their reliability depends on the quality, quantity, and coverage of labelled samples, thus expert time continues to be a constraint. Active learning (AL) eases this bottleneck by selecting, under a fixed annotation budget, the samples expected to improve a model most. The published evidence shows that AL can reduce the number of labels needed to reach a target level of predictive performance. Monitoring programmes, however, face a broader question: how should a limited expert budget be divided so that model training, model validation, and the ecological estimates built on model outputs all remain reliable? Because AL selects samples non-randomly, the labels it produces are unsuitable for validation, calibration, or threshold selection, and this tension is rarely acknowledged. We synthesise AL research across acoustic, image modalities, and identify several gaps and opportunities. Most published works evaluate query strategies on pre-labelled benchmark datasets with simulated annotators; deployments embedded in real monitoring workflows are rare and concentrate on birds and cetaceans. Bats, insects, amphibians, and fish are underrepresented, and multimodal applications remain largely unexplored. Evaluation practice centres on headline reductions in annotation effort, often without random-sampling baselines, per-class results, or calibration analysis, and almost never accounts for the labels required for validation. We provide a tutorial treatment of the AL loop that makes these budget decisions explicit, and a roadmap towards AL methods that support not only label-efficient training but also validation and trustworthy downstream ecological inference.

\end{abstract}


\section{Introduction}

\begin{figure}[H]
    \centering
    \includegraphics[width=0.83\linewidth]{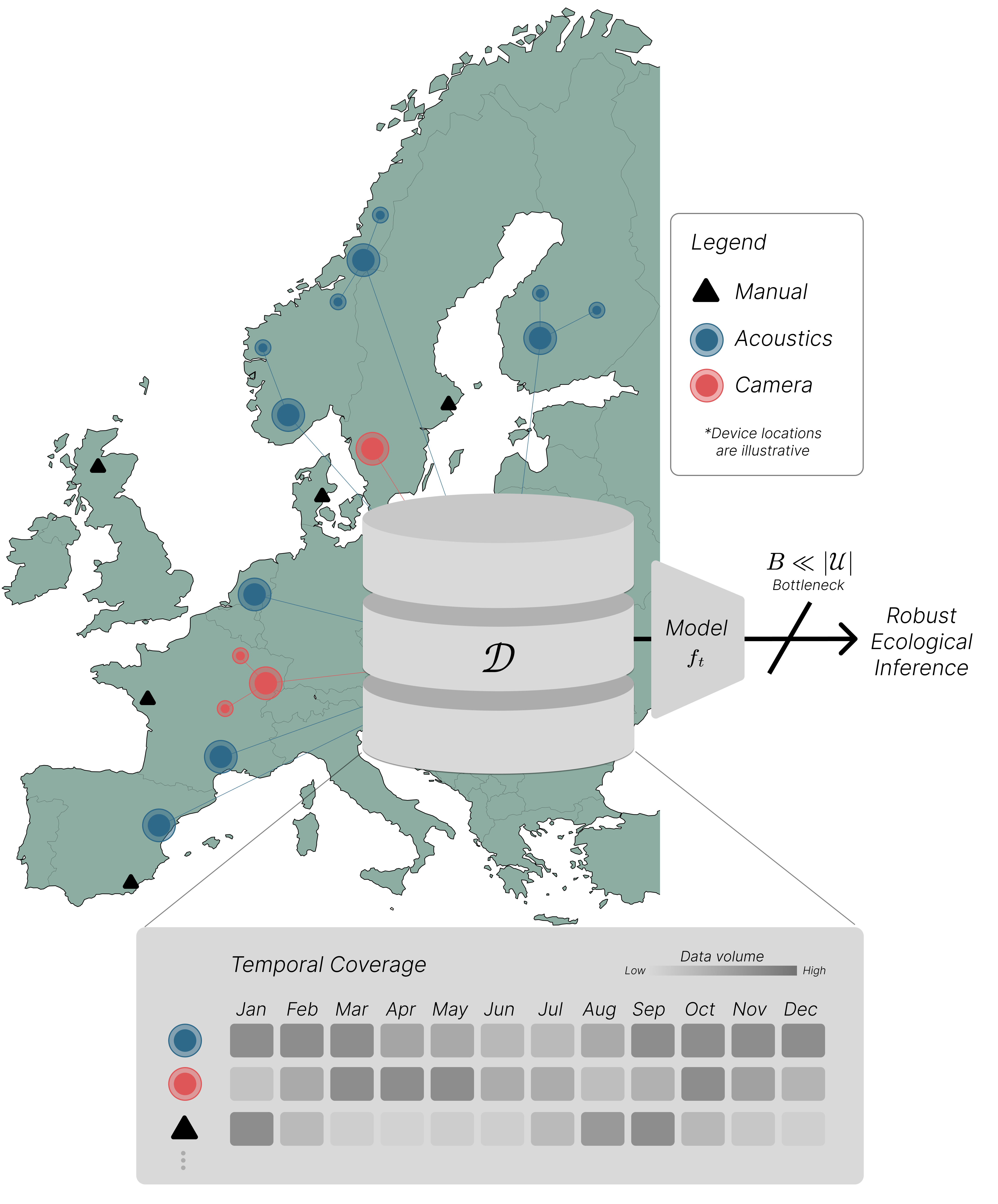}
    \caption{In-situ monitoring devices significantly increase the spatial and temporal resolution of monitoring initiatives yet they also increase data volumes. Ecological inference is constrained by both the information captured due to study design and the efficiency with which that information is extracted and validated.}
    \label{fig:bottleneck}
\end{figure}

Passive acoustic recorders and camera traps are expanding the spatial coverage and temporal extent of biodiversity monitoring. These approaches produce large volumes of recordings, images, and molecular observations, yet converting raw observations into species detections and ecological measures still depend heavily on expert annotation. Machine learning (ML) can process these data at scale, but model reliability depends on the quality, quantity, and coverage of labelled samples \citep{stowell2022computational, norouzzadeh2021deep, kitzes2025integrating}. Expert time therefore becomes a scarce resource in the monitoring workflow, and many projects can label only a small fraction of the data they collect \citep{kholghi2018active, norouzzadeh2021deep, van2023active} (Figure~\ref{fig:bottleneck}). This constraint creates a crucial allocation problem: \textit{when the pool of candidate data exceeds the annotation budget, which data is most valuable to label?}

\begin{figure}[H]
    \centering
    \includegraphics[width=0.75\linewidth]{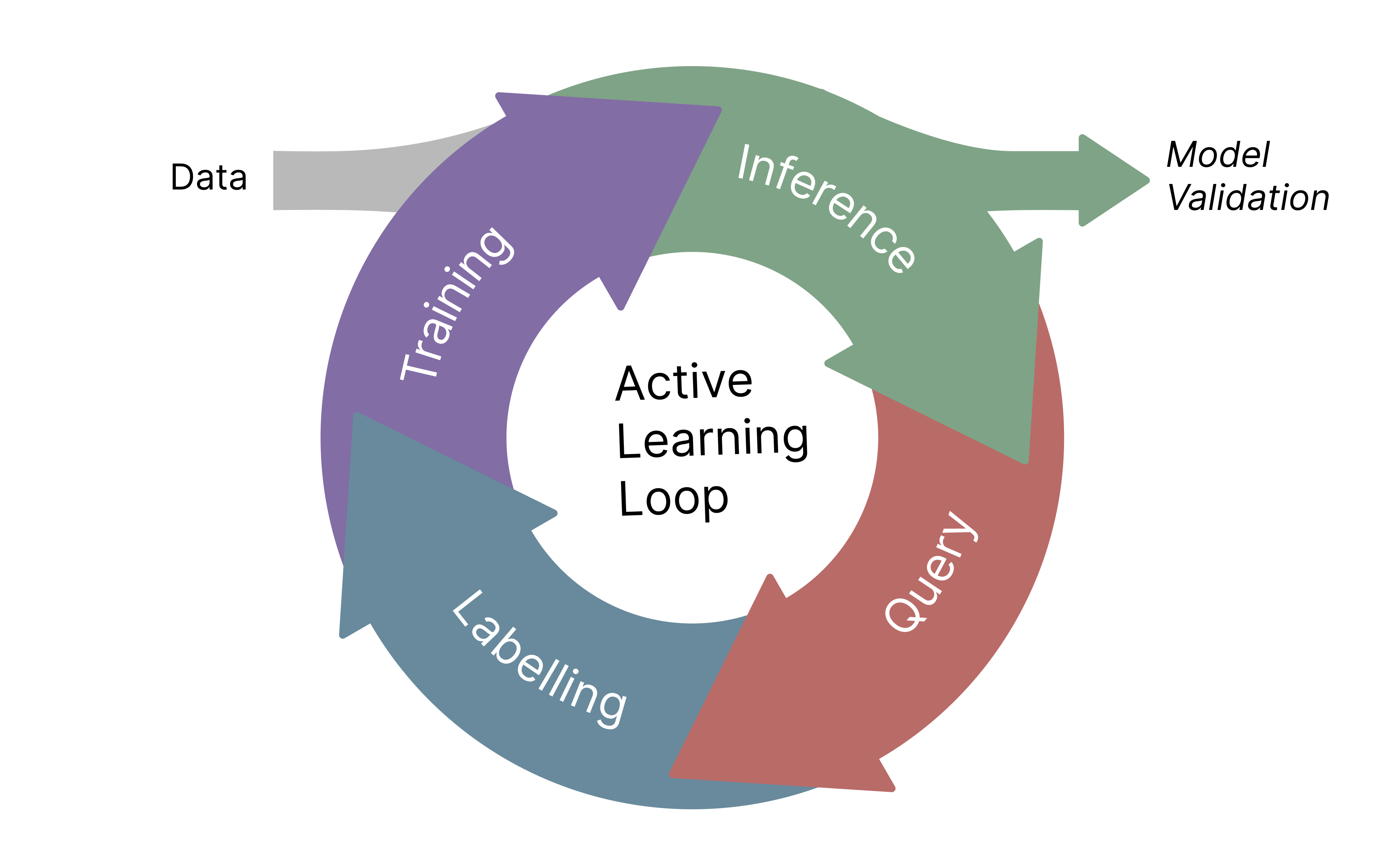}
    \caption{Overview of active learning pipeline.}
    \label{fig:intro}
\end{figure}

Active learning (AL) addresses this problem by using the current model and its representation of the data to guide subsequent sample selection \citep{settles2009active}. It aims to concentrate expert effort on samples expected to improve a specified learning objective, such as reducing classification error or improving recognition of rare classes \citep{mcewen2024active, kurinchi2026finding}. Annotation, model updating, and further querying form an iterative cycle (Figure~\ref{fig:intro}) in which the allocation of a limited annotation budget adapts to the model's needs. Studies in acoustic and image-based monitoring have used this approach to reduce the number of labels required to train species classifiers and detectors \citep{kholghi2018active, norouzzadeh2021deep, van2023active}. AL thus converts annotation from a one-off data preparation step into a iterative decision process within model development.

\begin{figure}[H]
    \centering
    \includegraphics[width=0.85\linewidth]{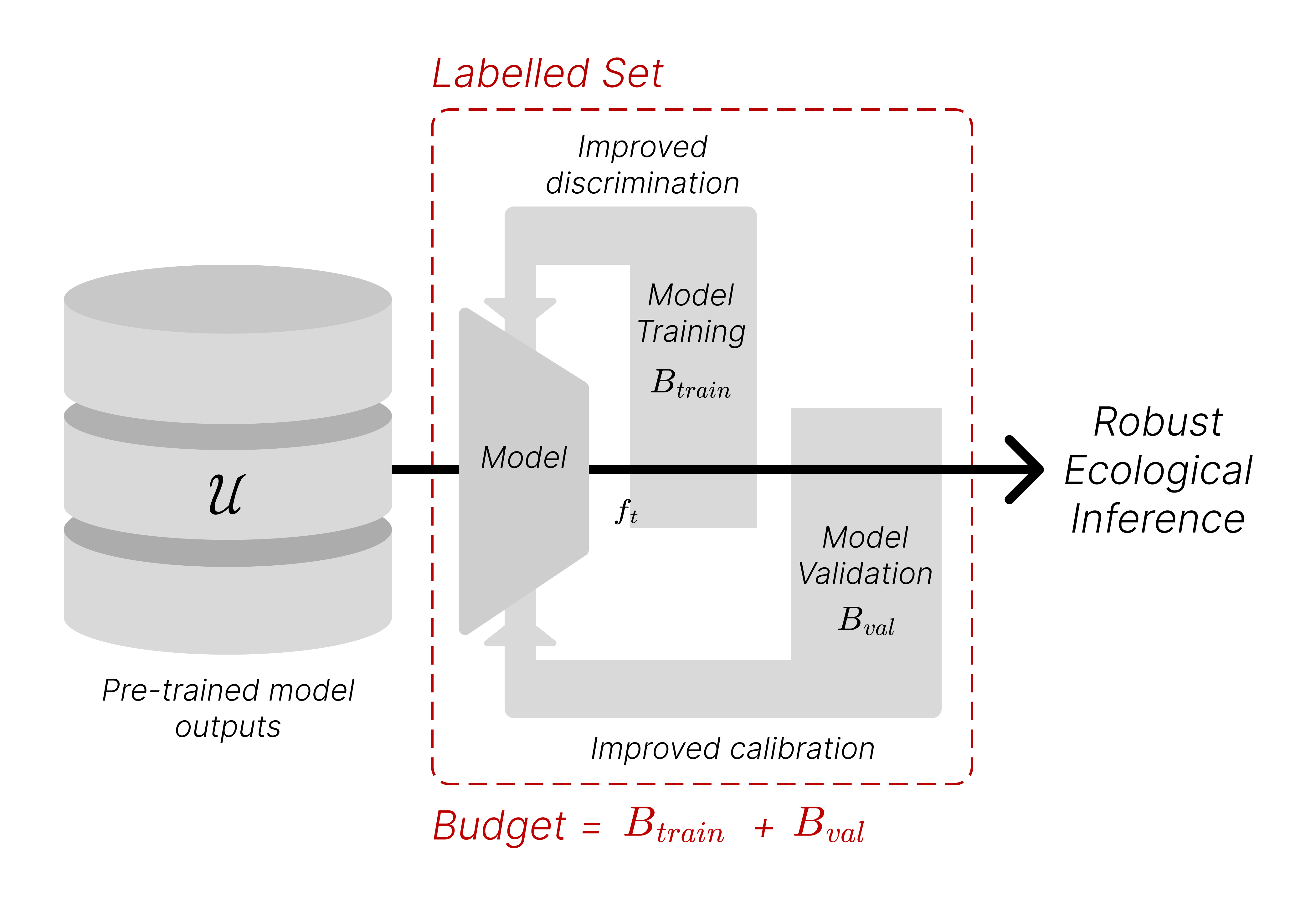}
    \caption{Model training and validation both draw from the same budget and are generally not interchangeable resulting in a dual allocation problem. \(\mathcal{U}\) denotes the pool of unlabelled samples, and \(B_{\mathrm{train}}\) and \(B_{\mathrm{val}}\) the shares of the expert budget spent on training and validation labels (Section~\ref{sec:budget}).}
    \label{fig:budget}
\end{figure}

Label efficiency, however, is not the only requirement that biodiversity monitoring places on annotation. Ecological data have long-tailed class distributions, strong spatial and temporal variation, heterogeneous observation conditions, and per-label annotation costs that vary widely. More fundamentally, the expert time that funds model training comes from the same budget as the expert time required for model validation and the verification of detections used for ecological estimates. A query strategy can improve classification performance while shifting the labelled sample away from the population distribution \citep{farquhar2021statistical}; labels acquired this way cannot be reused for validation, calibration or threshold selection without correction. Evaluating predictive performance at a given label count therefore says little about whether the occupancy, abundance, or phenology estimates built on the resulting model can be trusted. In short, existing biodiversity AL research largely demonstrates that fewer labels can suffice for predictive performance, whereas monitoring programmes need to know how a fixed expert budget should support reliable training, validation, and ecological inference together (Figure~\ref{fig:budget}). This review is organised around that gap.

Relevant studies are dispersed across acoustic and image modalities and across methodological communities, and in most cases evaluate query strategies on pre-labelled benchmark datasets with simulated oracle annotation. Ecological practitioners therefore find it difficult to judge whether a method fits a particular monitoring objective or to compare reported gains across studies. We synthesise 164 studies of active learning and machine learning for biodiversity monitoring. 

\subsection{Contributions}

This review makes three contributions.
\begin{itemize}
    \item A tutorial treatment of the active learning loop written for ecologists and field biologists, making explicit where each stage of the loop consumes expert effort and where design choices affect the reliability of later validation and ecological inference.
    \item A systematic map of 164 studies across acoustic and image modalities, showing that the literature concentrates on label-efficient training evaluated on pre-labelled benchmarks, and that validation practice, budget accounting, and downstream ecological objectives receive little attention.
    \item A roadmap for closing this gap, covering evaluation standards, explicit budget allocation between training and validation, cost-aware annotation, and the treatment of query-induced sampling bias in ecological inference.
\end{itemize}

The remainder of this paper is structured as follows. Section~\ref{sec:tutorial} introduces the AL loop in tutorial form and identifies where each stage draws on the expert budget. Section~\ref{sec:methodology} describes the survey method. Section~\ref{sec:map} maps the state of current literature. Section~\ref{sec:gap} analyses why label-efficient training alone does not deliver reliable monitoring and sets out a roadmap.

\section{A Budget-Aware Introduction to Active Learning}
\label{sec:tutorial}
This section provides a working introduction for readers who plan to use AL within a monitoring workflow. Unlike general treatments, we track a single quantity throughout: the expert budget. Every stage of the loop spends from it, and design choices at some stages determine whether the rest of the budget can be spent effectively. Readers seeking the general theory and method families of AL are referred to \citet{settles2009active} and \citet{ren2021survey}.

\subsection{Definitions}

A \textit{model} in this review is a function, generally a deep neural network, that maps an input (an audio segment or an image) to an output (predicted classes with confidence scores). Internally, the network converts the raw input into a set of numerical descriptors called \textit{features}. The vector of features produced by the final layers of the network is called an \textit{embedding}. Samples that lie close together in embedding space are treated as similar by the model, so embeddings provide a geometry over the data that many AL methods exploit. A \textit{pretrained model} has been trained beforehand on a large dataset, as with BirdNET \citep{kahl2021birdnet}. Its \textit{encoder}, the part that produces embeddings, can be \textit{frozen} (its weights held fixed) and used only to embed the data, with a lightweight classification head trained on top.

This review concerns deep learning: most of the surveyed studies use deep neural networks, and concepts such as embeddings and frozen encoders presuppose them. The general AL framework is also applicable to most ML methods.

\textit{Active learning} (AL) is an iterative process of model-guided sample selection and label acquisition, which aims to reach a specified learning objective with fewer labelled samples than passive (random) sampling \citep{settles2009active}. The most common objective is the discriminative performance of the model, but broader objectives such as annotation cost or calibration fit the same framework. The rule that scores and selects samples is the \textit{query strategy} (Section~\ref{sec:query-strategies}).

\textit{Label efficiency} is the number of labelled samples required to reach a specified level of performance, or equivalently the performance reached for a given number of labels. AL improves label efficiency when it reaches the target with fewer labels than random sampling. The learning curve (Section~\ref{sec:evaluation-practice}) is its standard visualisation.

The \textit{expert budget} is the total amount of expert annotation effort a monitoring programme can afford, counted in labels or in expert hours. Every stage of the AL loop draws on it.

\textit{Pool-based} AL assumes that a large collection of unlabelled samples, the pool, is available before annotation starts, as with an archive of recordings or camera-trap images. The query strategy scores the whole pool and selects from it. In \textit{stream-based} AL, samples arrive one at a time, for example on a recording device, and the decision whether to query each one is made on arrival. Almost all biodiversity AL is pool-based, matching the offline archives that passive sensors produce. A separate distinction concerns how many samples are queried per cycle: \textit{sequential} AL queries one sample and updates the model before the next query, whereas \textit{batch-mode} AL queries a batch of samples per cycle (Section~\ref{sec:loop}).

AL is frequently paired with \textit{human-in-the-loop} (HITL) workflows \citep{monarch2021human}. The two are related but distinct. HITL refers to any closed-loop process in which people participate in model development or verification, whereas AL refers specifically to model-guided sample selection.

AL as defined above selects labels for model \textit{training}. \textit{Active testing} applies the same idea to model \textit{validation}: it selects which samples an expert should verify so that model performance can be estimated to a target precision with fewer labels \citep{kossen2021active}. Both draw on the same expert budget, an allocation challenge discussed in Section~\ref{sec:budget}.

\subsection{One Expert Budget, Two Demands}
\label{sec:budget}

Biodiversity monitoring consumes expert effort at several stages of a pipeline: deploying sensors across ecological gradients, annotating detected signals, and validating and interpreting model outputs. This review concerns the latter two stages, which share one allocation problem: reduce uncertainty about a quantity of interest with as few expert observations as possible. Active learning addresses model training: it selects which samples to label so that the model improves most efficiently \citep{settles2009active}. Active testing addresses model validation: it selects which model outputs an expert should verify to characterise model performance most precisely under a fixed budget \citep{kossen2021active}, or stops verification once the confidence interval is acceptably narrow \citep{perez2024discount, perez2024human}. The two demands share one objective, maximising information per unit of expert effort, and they compete for the same budget. Writing \(B\) for the total number of expert labels (or expert hours) a programme can afford, \(B = B_{\mathrm{train}} + B_{\mathrm{val}}\), where \(B_{\mathrm{train}}\) is spent on training labels and \(B_{\mathrm{val}}\) on validation labels. How the budget is divided is itself a design decision, and it is the decision this review keeps returning to.

A related idea, \textit{adaptive sampling}, applies the same principle to field deployment itself, reallocating monitoring effort towards locations or periods where ecological utility is highest \citep{balantic2019temporally, alawad2022adaptive}. Adaptive sampling is absent from the reports and deployments we survey, and requires specific circumstances such as mobile monitoring tools. We therefore leave it outside the scope of this review, noting only that adaptive sampling would place additional effects on the same expert budget.

The budget question extends beyond which samples to label for training. A query strategy selects a sample precisely because it is unusual under the current model, so the labelled set it produces is not a random sample of the monitored population \citep{farquhar2021statistical}. Uncertainty sampling, for example, concentrates on decision-boundary cases: error rates estimated on such a set are systematically pessimistic, and calibration curves or detection thresholds fitted to it do not transfer to the full data stream. Labels bought through active learning therefore serve training well but validation poorly, and the two purposes require separately budgeted data \citep{lowell2019practical, kitzes2025integrating}.

The distinction matters most when model outputs feed ecological analyses. Occupancy, abundance, and phenology estimates inherit the error characteristics of the detections they are built on, and therefore require validated error rates across the relevant sites, seasons, and classes rather than a single global accuracy figure \citep{kitzes2025integrating}. Throughout this review we accordingly apply a simple test: \textbf{an AL study is \textit{budget-complete} only if it accounts for \(B_{\mathrm{val}}\)}, the labels needed to validate the resulting model, alongside \(B_{\mathrm{train}}\). Section~\ref{sec:gap} returns to the mechanics of query-induced bias and to methods for validating under a shared budget.

\subsection{The Active Learning Loop}
\label{sec:loop}
\begin{figure}[H]
    \centering
    \includegraphics[width=0.80\linewidth]{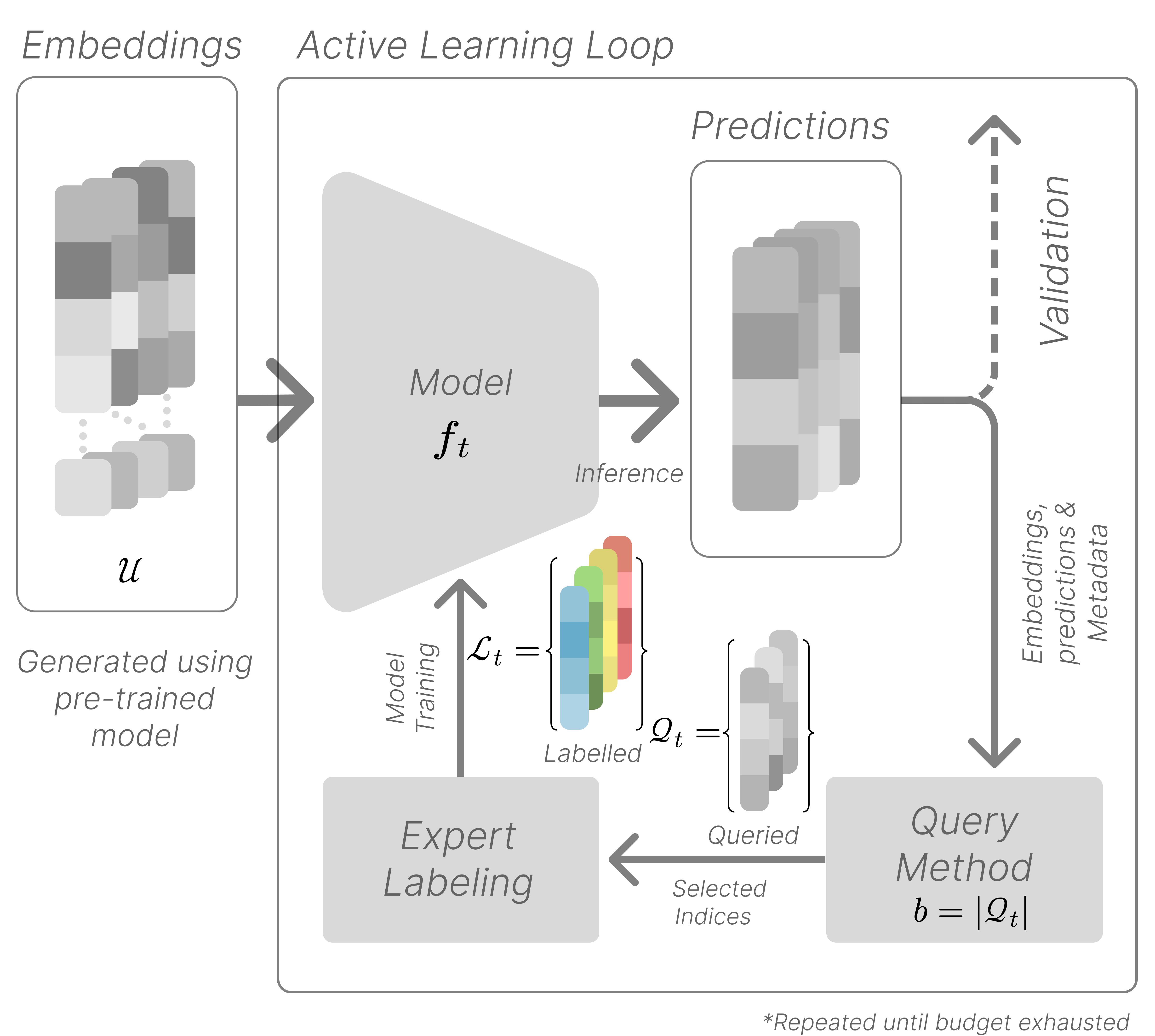}
    \caption{Active learning cycle showing active fine-tuning of model, inference, querying and expert labelling. At cycle \(t\), the model trained on the labelled set \(\mathcal{L}_t\) is run over the unlabelled pool \(\mathcal{U}\), and the query method selects a batch \(\mathcal{Q}_t\) of \(b\) samples for expert labelling.}
    \label{fig:active-learning}
\end{figure}

A full loop consists of initialisation, inference, querying, annotation, training, and a stopping criterion (Figure~\ref{fig:active-learning}). We describe each stage in turn and note where it draws on the expert budget. In our notation, \(f_t\) denotes the model at cycle \(t\), \(\mathcal{U}\) the pool of unlabelled samples, \(\mathcal{L}_t\) the accumulated labelled set on which \(f_t\) is trained, \(\mathcal{Q}_t \subset \mathcal{U}\) the batch queried in cycle \(t\), and \(b = |\mathcal{Q}_t|\) the batch size.

\paragraph*{Initialisation.} Sample selection is guided by model outputs, so the quality of selection depends on the discriminative capability of the initial model \(f_0\). For an uninitialised model, early selections can be poor (the ``cold-start problem''). The initialisation (warm-up) phase typically is the first phase of training on a small labelled set \(\mathcal{L}_0\), which is generally randomly sampled \citep{lindholm2025aggregation, norouzzadeh2021deep} or may already exist.\\

It has become increasingly common to use a frozen pretrained encoder and apply AL to the training of a lightweight classification head (Section~\ref{sec:models}). In this case, other initialisation methods can diversify sampling across the embedding space, such as Core-Set (farthest-first traversal) \citep{sener2018active}, sampling of representative high-density regions such as TypiClust \citep{hacohen2022typiclust}, compared on bioacoustic datasets by \citet{rauch2024towards}, or k-means++ \citep{arthur2007k}, applied to aquatic invasive species recognition by \citet{chowdhury2024active}. When labelled samples are present but very scarce, distance-based sampling can be applied to prototypical class embeddings (the mean of a class support set) \citep{mcewen2024active}. With increasingly capable pretrained models it is also common to skip initialisation and apply informativeness-based or hybrid methods directly. Clearer guidance on when initialisation utility and method are needed. A reasonable default is that a pretrained encoder covering the target classes removes the need for a separate warm-up \citep{tamkin2022active}.

\paragraph*{Inference.} The current model \(f_t\) is run as a single forward pass over the unlabelled pool (e.g.\ audio segments or images), producing for each sample a predicted class, a confidence score, and an embedding. The model's confidence scores and embeddings are then used to inform the query strategy.

\paragraph*{Querying.} The query strategy selects a subset \(\mathcal{Q}_t\) of the unlabelled pool to present to the annotator, ranked by estimated utility. Utility is typically decomposed into two complementary properties: \textit{informativeness}, the expected reduction in model uncertainty if the sample were labelled, and \textit{non-redundancy}, the degree to which a sample adds information not already represented by the other samples selected in the same cycle.\\

In sequential AL, one sample is queried per cycle (\(b = 1\)) and the model is updated before the next query. This maximises informativeness but is computationally prohibitive at the scales typical of passive biodiversity monitoring, both because of the volume of data collected and because of the infrastructural difficulty of closing the loop between querying, model updates, and human annotators. Sequential AL is therefore uncommon in the biodiversity literature and viable only in simulation, through oracle sampling\footnote{Oracle labelling is the process of revealling pre-labelled samples, simulating the querying and expert annotation process.} of pre-labelled data.\\

Batch-mode AL, which selects \(b > 1\) samples per cycle, is more commonly applied. The model is updated once after the batch is annotated, trading sample efficiency for computational efficiency. Note the batch size \(b\) refers to the number of samples selected per cycle, \textit{not} the batch size used for model training.
Although performance generally degrades at larger batch sizes \citep{citovsky2021batch, zhang2025evaluating}, the batch size is more often set by feasibility constraints or simple heuristics. Common batch sizes range from 20 \citep{mcewen2024active, kath2024active} to 100 \citep{norouzzadeh2021deep, qian2017active}. Some studies use larger batches, such as the 1,712 images per cycle of \citet{nguyen2025model}, and work outside biodiversity extends to extreme batches of 100k to 1M images \citep{citovsky2021batch}. Ablation studies exploring batch size are limited \citep{zhang2025evaluating}. \citet{mononen2025active} set the batch size from the number of predicted classes (50 samples per class). Evidence from the general AL literature suggests that dynamic batch sizes and batch scheduling, starting small and growing, could improve performance \citep{citovsky2021batch}, but this remains largely unexplored. Given the resource and computational constraints of biodiversity monitoring, batch size and large-batch AL merit further investigation.

Batch mode matches offline, pool-based annotation workflows but introduces a redundancy problem: samples selected independently by an uncertainty criterion may be near-duplicates, wasting annotation budget \citep{citovsky2021batch}. Diversity-aware strategies (Section~\ref{sec:query-strategies}) address this by enforcing coverage of the unlabelled pool within each batch.

\paragraph*{Annotation.} The selected samples \(\mathcal{Q}_t\) are presented to a domain expert, typically a trained ecologist or taxon specialist, who provides labels, so each cycle spends \(b\) labels of \(B_{\mathrm{train}}\). These labels are generally treated as ground truth, although inter-annotator disagreement is sometimes considered \citep{van2023active}. The per-sample annotation burden in biodiversity monitoring varies substantially: a clearly vocalising target species takes seconds to confirm, while an ambiguous vocalisation, a rare species, or an occluded image may require extensive analysis. Ambiguous samples sometimes need to be skipped \citep{mononen2025active} or labelled as uncertain \citep{van2023active}. Samples acquired to optimise model sample efficiency can therefore come at the cost of annotation efficiency, since query strategies often select ambiguous cases.

The annotation task itself also varies in form and duration: confirming a species-level prediction is simpler than drawing a bounding box on an image or spectrogram, and providing strong multi-label annotations is more cumbersome than weaker single-label annotations \citep{martinsson2025accuracy}. This variability motivates methods that reduce the cost of each label independently of sample selection, such as model-guided weak labels \citep{martinsson2024weak}; we return to these in Section~\ref{sec:active-annotation}.

In AL benchmarking and method development, the annotation process is simulated through \textit{oracle labelling}: AL runs on a pre-labelled dataset and labels are revealed after querying. As a consequence, under oracle labelling, annotation cost is largely ignored.

\paragraph*{Training.} Following annotation, the queried batch joins the labelled set, \(\mathcal{L}_{t+1} = \mathcal{L}_t \cup \mathcal{Q}_t\), and the model is fine-tuned on the full accumulated labelled set to give \(f_{t+1}\). When only a classification head is updated on a frozen backbone, the embedding space remains fixed across cycles. When the full network is fine-tuned \citep{mohaimenuzzaman2023deep, chowdhury2024active}, training and querying become directly coupled: as model weights update, the embedding space shifts, reshaping what the query strategy selects in the next cycle. \citet{mohaimenuzzaman2023deep} demonstrate this in bioacoustic AL, showing that updating the feature extractor within the loop outperforms static-feature approaches, and attribute the gain to the representation adapting to the newly labelled distribution. \citet{tamkin2022active} show that AL effectiveness is an emergent property of representation quality: pretrained models with linearly separable feature spaces require up to five times fewer labels, and hence a lower \(B_{\mathrm{train}}\), than their non-pretrained counterparts.

\paragraph*{Stopping criterion.} The annotation budget defines the maximum annotation effort available across all AL cycles; in biodiversity monitoring it is typically constrained by expert availability. Reported budgets vary considerably, from a few hundred samples to tens of thousands for large camera-trap datasets \citep{norouzzadeh2021deep}, reflecting the diversity of task difficulty, pool size, and deployment context. Cycles terminate when the budget is exhausted, when a fixed number of iterations is reached \citep{chowdhury2024active}, or when a performance threshold is met \citep{williams2025evidence}. The last approach is preferable in principle because it adapts to the difficulty of the problem, but it requires a held-out validation set to monitor performance. When labels are scarce, partitioning them for validation is costly; \citet{tamkin2022active} propose stopping when the training loss falls to a fixed fraction of its initial value, avoiding a separate validation set. In practice, stopping criteria are rarely reported explicitly in the biodiversity AL literature: most studies run until a fixed budget is exhausted or report results at a predetermined number of cycles, which makes cross-study comparison of label efficiency difficult. Notably, a performance-based stopping criterion is the first point inside the loop where validation labels must be spent.\\

\subsection{Query Strategy Families}
\label{sec:query-strategies}
Query strategies can be grouped into three families by the signal used to rank samples. \textit{Uncertainty-based} strategies use the current model's predictions: least-confidence, margin, and entropy criteria rely on class probabilities, while ensemble or Bayesian approximations additionally estimate model (epistemic) uncertainty at a higher computational cost \citep{settles2009active, ren2021survey}. Uncertainty sampling concentrates queries near decision boundaries, which is efficient for refining a classifier but is also the source of the sampling bias discussed in Section~\ref{sec:gap}.

\textit{Geometry-based} strategies use only the structure of the embedding space: diversity methods such as Core-Set maximise the distance between samples within a batch \citep{sener2018active}; coverage methods ensure that all regions of the pool are represented; and typicality methods such as TypiClust prioritise high-density regions \citep{hacohen2022typiclust}. Because these strategies do not depend on reliable class probabilities, they are particularly useful in cold-start and low-budget settings.

\textit{Hybrid} strategies combine informativeness with diversity and are the default choice in batch-mode settings because they suppress within-batch redundancy \citep{citovsky2021batch, zhang2025hybrid, kath2024active}. Beyond these three families, the ranking objective itself can change: cost-aware strategies include the expected annotation cost of each sample in its utility, and ecological-utility strategies select samples by the expected reduction in uncertainty of a downstream ecological quantity rather than model accuracy, for example sampling in service of species distribution models \citep{lange2023active} or stratification across spatial, temporal, and ecological gradients \citep{mcewen2025stratified}. This last class of objectives is the most directly relevant to biodiversity monitoring and the least studied.

\section{Review Methodology}
\label{sec:methodology}

\subsection{Survey design}

\begin{figure}[H]
    \centering
    \includegraphics[width=0.8\linewidth]{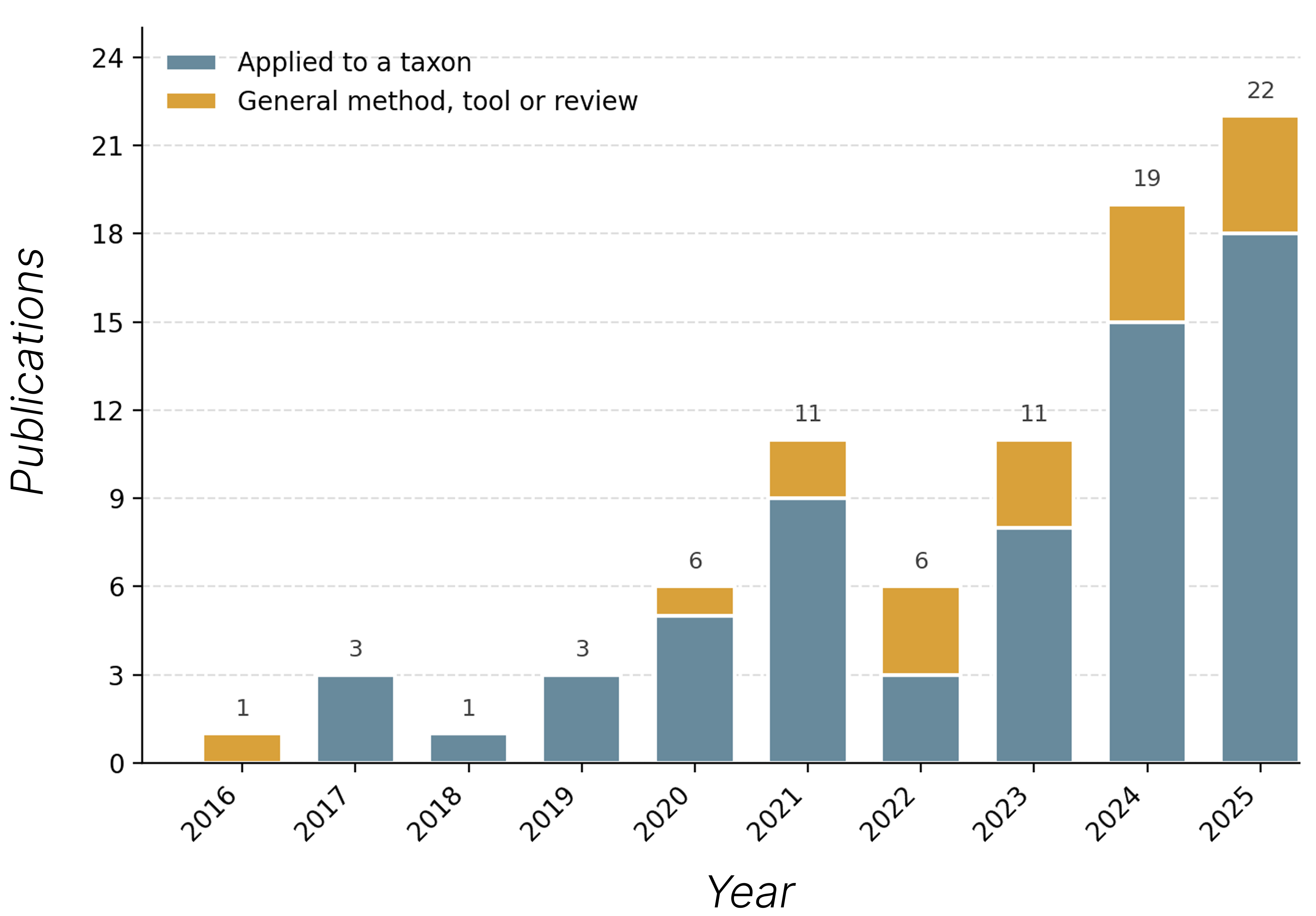}
    \caption{Number of published papers related to active learning and biodiversity monitoring over time.}
    \label{fig: publishing}
\end{figure}

An initial set of searches was conducted on the 27th of November 2025 using Google Scholar, covering the passive monitoring modalities relevant to biodiversity monitoring (acoustics and camera traps). The search queries for each modality are listed below. No date range was specified. The number of relevant papers and pre-prints increases over time (Figure~\ref{fig: publishing}).

\paragraph*{Bioacoustics:} 553 search results with \textbf{192 selected} provided by the following query:
("active learning" OR "query learning" OR "selective sampling" OR "human-in-the-loop") AND (bioacoust* OR ecoacoust* OR vocali* OR “animal calls” OR "passive acoustic monitoring" OR "soundscape") AND (animal OR bird* OR cetacean* OR insect* OR mammal*)

\paragraph*{Camera trap:} 571 search results with \textbf{171 selected} provided by the following query:
("active learning" OR "query learning" OR "selective sampling" OR "human-in-the-loop") AND ("camera trap*") AND (animal* OR wildlife OR mammal* OR bird* OR "biodiversity monitoring")

The searches returned \textbf{1128 results} in total. After a preliminary review of abstracts, \textbf{367 papers} were selected for further review and exported from Google Scholar. After deduplication and removal of irrelevant literature (books, theses and inaccessible papers), \textbf{156 papers} entered comprehensive review. A supplementary search was conducted on the 21st of April 2026 using Claude Opus 4.6 (in research mode) with the Consensus AI and BioRxiv connectors enabled, prompted with the existing queries and paper list; this identified \textbf{8 additional relevant papers}. The final corpus comprises \textbf{164 papers}.

\paragraph*{Schema:} Following the identification of relevant literature, a schema of 24 questions across seven categories was developed: paper metadata (e.g.\ preprint status), AL configuration (e.g.\ query strategy, budget), performance and evaluation (e.g.\ evaluation metrics), application domain (e.g.\ modality, taxonomic coverage), technical details (e.g.\ base model, datasets), context (e.g.\ motivation and use case), and summary. The full schema is provided (Appendices~\ref{app:schema}). The schema gave a consistent framework for evaluating the reviewed papers. Generative AI (Claude) was used to aid the extraction of schema information; all information reported in this review has been manually reviewed and verified by the authors.

\section{The State of the Field}
\label{sec:map}
We organise the literature review around four questions: which data and taxa are studied (Section~\ref{sec:coverage}), which models the loop is built on (Section~\ref{sec:models}), which query strategies are used in practice (Section~\ref{sec:strategies-in-use}), and how success is evaluated and supported by tools (Sections~\ref{sec:evaluation-practice} and~\ref{sec:tools}). The answers to all four point to the same conclusion: the experimental designs of the current literature demonstrate label-efficient training, rather than the budget-complete validation and inference that monitoring and ecological inference requires.

\subsection{Modalities, Taxa, and Datasets}
\label{sec:coverage}
The current literature is heavily weighted toward methods papers that reuse existing ecological datasets (BirdSet \cite{rauch2025birdset}, AudioSet \cite{gemmeke2017audio}, Snapshot Serengeti \cite{swanson2015snapshot}, AnuraSet \cite{canas2023dataset}). A small number of corpora recur, with different query strategies evaluated on the same underlying recordings. This reuse, while valuable for the development and benchmarking of AL methods, does inflate taxonomic coverage. We therefore distinguish throughout between benchmark coverage, where a taxon is present only because it is contained in a reused dataset, and deployment coverage, where active learning is embedded in a workflow that answers a question about a real population. Deployment is the rarer case: 20 of the 66 taxon-related studies (30\%) report an applied deployment rather than a method or tool demonstration. These deployments are concentrated in terrestrial mammals (9 studies, mostly conservation camera-trap programmes for endangered species) and birds (7), followed by marine mammals (4) and fish (3).

\begin{figure}[H]
    \centering
    \includegraphics[width=\linewidth]{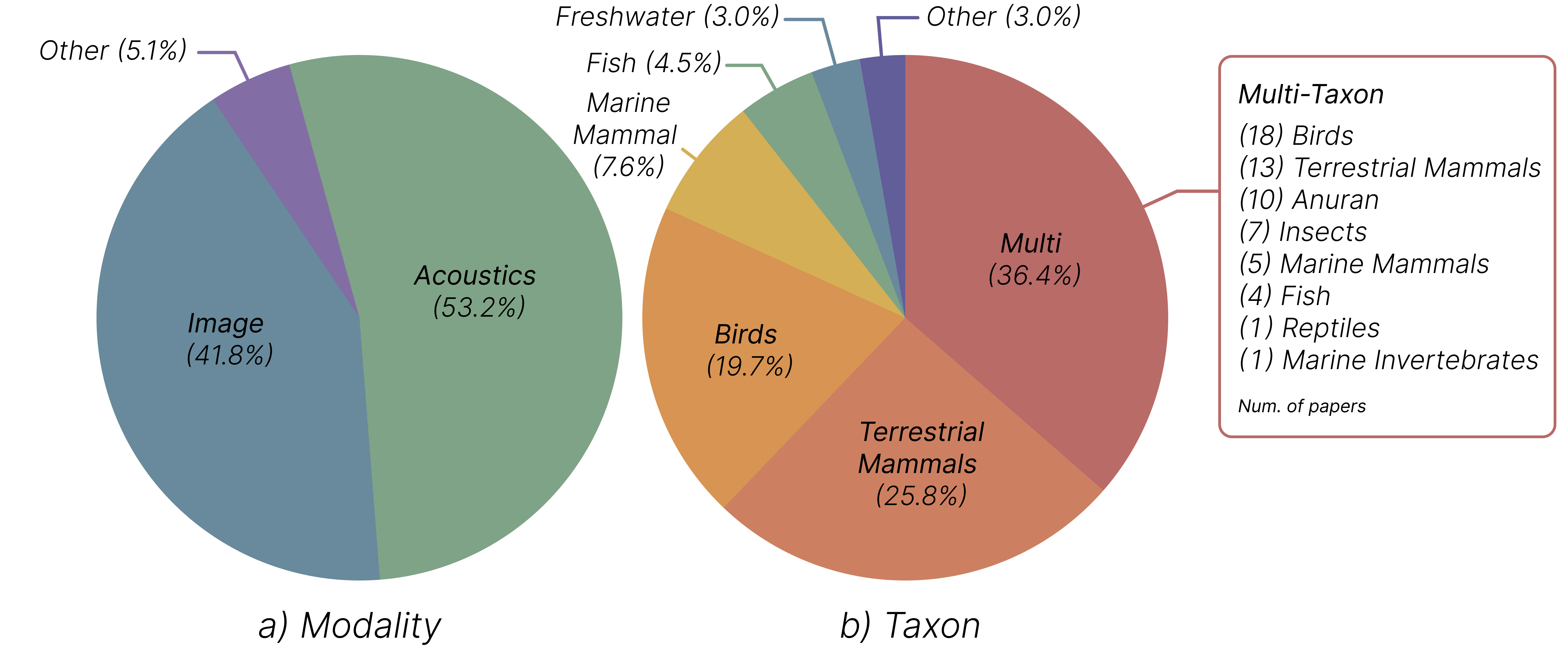}
    \caption{a) Data modality split across active learning literature and b) taxonomic split across 66 taxon-specific active learning papers.}
    \label{fig:modality-taxa}
\end{figure}

\paragraph*{Taxonomic Coverage}
Assigning each of the 66 taxon-related studies, the largest group is multi-taxon (35\%), which are evaluated on multi-species benchmarks. Terrestrial mammals follow at 26\% and birds at 20\%, then marine mammals (8\%), fish (5\%), freshwater species (3\%), and the remaining studies focus on marine invertebrates and plants. Bats are absent from AL literature. Insects and anurans are almost entirely confined to the multi-taxon studies: insects are the dominant taxon in no study despite appearing in seven, and work on anurans is entirely constrained to AnuraSet \citep{canas2023dataset}.

Birds are the most-investigated taxon and the one with the clearest record of field deployment. On the methods side, \cite{rauch2024towards} benchmark entropy-based and hybrid query strategies on BirdSet, and \cite{qian2017active} compare uncertainty- and diversity-based selection across sixty species. Deployments pair strategies with monitoring targets: \cite{mcewen2025stratified} apply stratified uncertainty sampling over BirdNET embeddings in the transnational TABMON network \cite{cretois2026tabmon}, \cite{ayers2021reducing} prioritise recordings for more than 150 threatened coastal species, \cite{bellafkir2023edge} drive edge-deployed recognisers of 214 species using ensemble reliability scores, \cite{van2023active} develop a recogniser for the endangered southern black-throated finch.

Terrestrial mammal studies occupy a similar share of the corpus but with a different emphasis. Much of the volume is benchmark-driven, with \cite{norouzzadeh2021deep} comparing query strategies on Snapshot Serengeti and NACTI and \cite{mononen2025active} applying core-selection to global camera-trap data, while genuine deployments pair strategies with sites, including \cite{miao2021iterative} at Gorongosa National Park and \cite{bothmann2023automated} and \cite{auer2021minimizing} in the Bavarian Forest. A distinct sub-strand applies active learning to individual re-identification \cite{kulits2021elephantbook, brust2020active, sani2025active}. The apparent richness of mammalian coverage is concentrated in a handful of reused megafauna and savanna-community datasets rather than distributed across mammalian monitoring contexts.

Coverage of the remaining groups is thinner and, in several cases, an artefact of dataset reuse. Amphibian coverage is entirely attributable to AnuraSet, which supplies the frog and toad classes evaluated by \cite{kath2024active} and \cite{kath2024leveraging} and appears as a testbed in generalisation studies \citep{mcewen2025stratified}; we found no study in which active learning was deployed to monitor a real anuran population. Marine mammals are the second group (after birds) with a substantive deployment record, through humpback whale song \citep{allen2021convolutional}, seasonal Bryde's whale calls \citep{allen2024bryde}, and blue and fin whale call-density estimation \citep{alksne2026application}, with benchmark work on the BioSED pilot-whale task \citep{zhang2025hybrid}. Insect coverage is incidental, arising through the BioSED mosquito task \citep{zhang2025hybrid}, cicada and cricket sounds in long-duration soundscapes \citep{kholghi2018active}, and image-based work on bees \citep{boinski2020collaborative}, even though reviews of emerging monitoring technology repeatedly flag insects as a priority \citep{van2022emerging, sheard2024emerging}. Fish coverage is also low \citep{bordoux2026rapid, williams2025evidence, dumoulin2025search}. Freshwater systems, which we separate from marine ones because their monitoring conditions differ, amount to two studies: semi-automated selection for endangered river dolphins in the Brazilian Amazon \citep{erbs2023towards}, the only freshwater deployment in the corpus, and contrastive representation learning with k-means selection for invasive dreissenid mussel larvae \citep{chowdhury2024active}. Reptiles appear only as minor classes within multi-taxa camera-trap datasets, and no study takes them as a target. Bats are absent, which is notable given that they are among the most intensively acoustically monitored taxa.

\paragraph*{Modalities}
Acoustic data accounts for 53\% of the studies with an identifiable sensing modality and camera-trap or other image data for 42\%, leaving 5\% across everything else (Figure~\ref{fig:modality-taxa}). Those remaining studies show active learning reaching sensing contexts the field has otherwise left alone. \cite{kang2021exploiting} exploit temporal and spatial proximity structure to surface rare hummingbird events in a 9{,}000\,GB field video archive. \cite{kozlova2025dlc2action} apply Bayesian active learning by disagreement to behavioural segmentation from pose and video features, shifting the annotation target from species identity to behaviour. \cite{lange2023active} estimate species ranges from geographic coordinates and environmental covariates, using a weighted committee of pre-trained per-species models to select where to sample next across a thousand mixed taxa. \cite{tamkin2022active}, counted here as image work, applies one selection method to both camera-trap imagery and NLP tasks. \cite{van2025arise} is the only multi-sensor entry, combining wildlife cameras, acoustic recorders, insect cameras and radar in a national monitoring infrastructure, though it describes interactive training cycles as an intended capability rather than reporting a query strategy in use.

\subsection{Models}
\label{sec:models}

Deep neural networks are the base model in over 94\% of the surveyed studies. In much of the recent bioacoustic and camera-trap literature, the starting point has shifted from training task-specific classifiers from scratch towards adapting pretrained encoders, detectors, or classifiers. The governing question is increasingly how to adapt a pretrained model or its classification head to a new domain (for example, a new location or recording condition) or a new application \citep{dumoulin2025search}. In bioacoustics, commonly used base models include BirdNET \citep{kahl2021birdnet}, the Perch family of birdsong embeddings \citep{ghani2023global, burnsperch}, self-supervised animal vocalisation encoders such as AVES and its avian variant BirdAVES \citep{hagiwara2023aves}, animal2vec \citep{schafer2024animal2vec}, masked-autoencoder models adapted to audio \citep{rauch2025can}, and general-purpose audio encoders such as BEATs \citep{chen2022beats}; a recent comparative review contrasts these models \citep{schwinger2025foundation}. For image-based camera-trap monitoring, MegaDetector provides a widely used animal, person, and vehicle detector \citep{beery2019efficient, hernandez2024pytorchwildlife}, complemented by visual foundation models such as BioCLIP \citep{stevens2024bioclip} and multimodal models for zero-shot species recognition \citep{fabian2023multimodal}. These systems differ in role, as classifiers, embedding models, or detectors, but all provide reusable predictions or representations. Pretrained embeddings can transfer across locations, species, and in some cases taxa, making them an effective basis for downstream tasks under limited supervision \citep{ghani2023global, ghani2025impact, williams2025using, burnsperch}.

Two recurring technical routes are visible in the literature. The first is end-to-end active learning, in which the full network is trained or fine-tuned within the loop \citep{rauch2023active}. The second uses a frozen pretrained encoder to embed the candidate pool and applies AL to a lightweight classification head fitted on the resulting embeddings \citep{rauch2024towards, kath2024leveraging, kath2024active_socialgood, kath2025speeding, zhang2025hybrid, lindholm2025aggregation, bernard2025data}. The frozen-encoder route is increasingly common: for a fixed candidate pool, embeddings can be cached, reducing the cost of repeated encoder inference. The acquisition step and classifier retraining can still be expensive for very large pools, but the approach is complementary with transfer learning because both reduce the quantity of labelled data required. Embeddings and predictions generated by pretrained models have therefore become an important starting point for AL in this field.

When AL is built on a pretrained model, the task becomes selecting, from a large unlabelled pool, the most informative subset to annotate for adapting that pretrained model under a fixed budget \citep{xie2023active}. The encoder may be frozen at its initial values or updated together with the classification head, but when the labelled seed set is absent or very small, the uncertainty signals that drive pool-based AL are weak. Cold-start selection therefore often relies on the geometry of the embedding space through coverage, diversity, and typicality criteria before uncertainty-based or hybrid strategies become reliable \citep{sener2018active, hacohen2022typiclust}. Related work reframes efficient AL through proxies and feature alignment in the pretrained-model setting \citep{wen2024feature} and examines how AL interacts with pretrained models \citep{tamkin2022active, bodesheim2022pre}. As pretrained models become a common substrate for biodiversity monitoring, the open problem shifts in part from which classifier to train towards which data to fine-tune on. Downstream performance is sensitive to the choice of embedding \citep{dumoulin2025search}, yet principled criteria for selecting an embedding to drive AL remain to be established; we identify this as an open question.

\subsection{Query Strategies in Use}
\label{sec:strategies-in-use}

In applied studies, uncertainty sampling is the default method for selecting samples \citep{qian2017active, kholghi2018active, norouzzadeh2021deep}: it is simple to implement and requires only model confidences as input. It carries a hidden cost, however: entropy- and confidence-based criteria tend to select acoustically or visually confusing samples, and precisely these samples are the hardest to annotate, sometimes having to be skipped or labelled as uncertain \citep{mononen2025active, van2023active}. Maximising the expected information per label can therefore simultaneously maximise the annotation cost per label.

A clear pattern is that monitoring applications benefit from hybrid methods that combine high-utility sampling with diversification. There are two reasons. First, batch sizes in monitoring are large, so the implicit diversification provided by sequential or small-batch settings is unavailable \citep{citovsky2021batch}. Second, as noted above, pure uncertainty criteria accumulate near-duplicate, ambiguous samples within a batch. Strategy comparisons on biodiversity data, with benchmarks on BirdSet, BioSED, and AnuraSet, support this observation \citep{rauch2024towards, zhang2025hybrid, kath2024active}.

A third usage pattern falls outside the textbook, model training usecase but recurs in practice: exploratory sampling, in which the model guides experts to the target events themselves within a large volume of data, such as calls of rare or invasive species, or hard negatives \citep{mcewen2024active, kather2024development, van2024seasonal, alksne2026application}. Related as well are vector-based similarity searches which is one component within the "agile modelling" workflow \citep{dumoulin2025search}. The goal is not to improve a classifier evenly but to confirm as many target detections as possible in limited time. This pattern is closest to the real motivation of many monitoring programmes, and it again illustrates that improvement in model accuracy is not the only quantity practitioners care about.

\subsection{Evaluation Practice}
\label{sec:evaluation-practice}
\begin{figure}[H]
    \centering
    \includegraphics[width=0.60\linewidth]{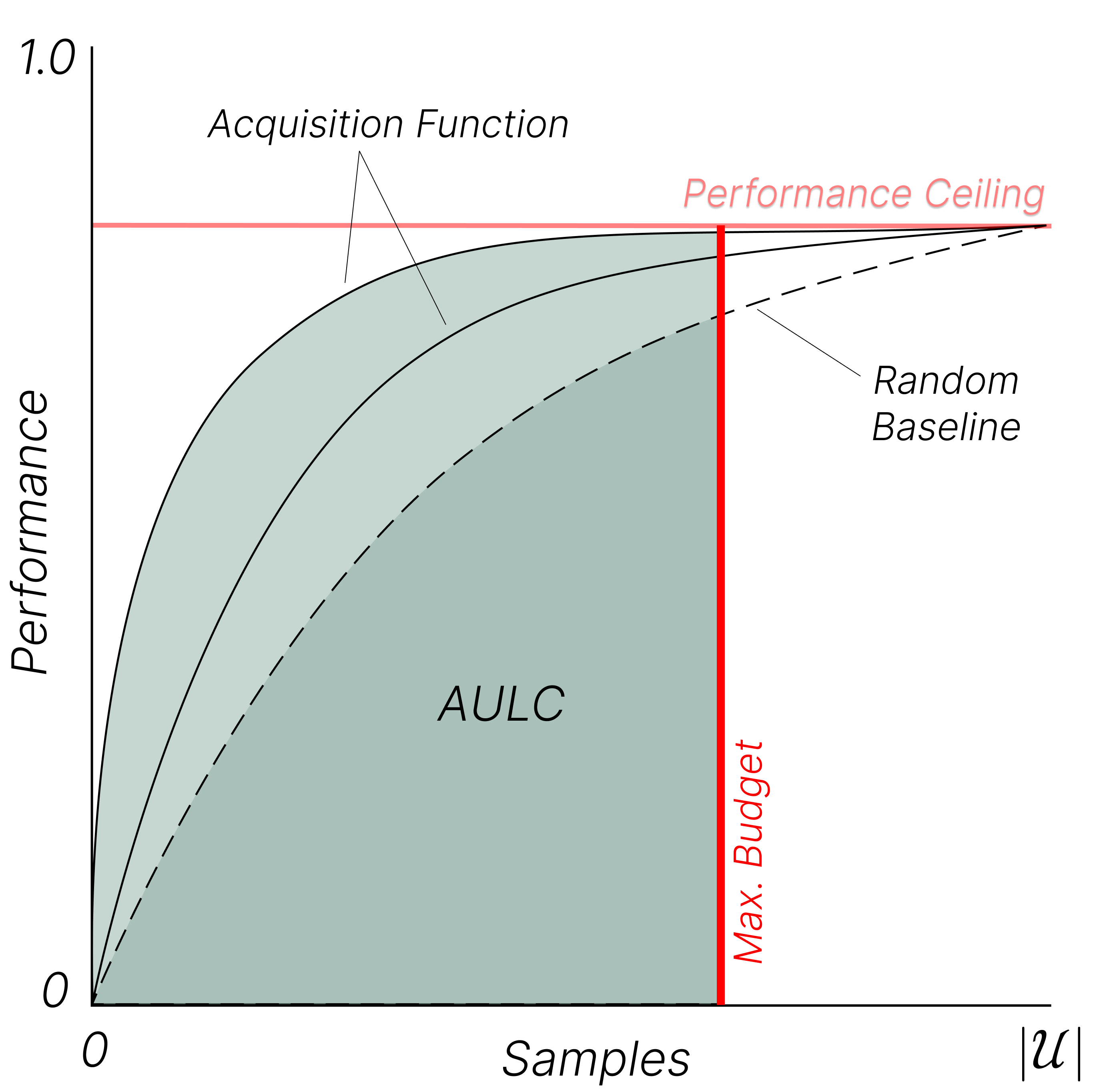}
    \caption{Active passive learning curves, showing performance ceiling and Area Under the Learning Curve (AULC) for a fixed budget. The horizontal axis is the number of labelled samples \(|\mathcal{L}_t|\), which reaches the pool size \(|\mathcal{U}|\) once the whole pool is labelled.}
    \label{fig:learningcurve}
\end{figure}

The dominant evaluation protocol reports a \textit{learning curve}: a performance metric measured as a function of the annotation budget (the number or proportion of labelled samples) (Figure~\ref{fig:learningcurve}). When a passive baseline is included, the AL strategy is compared against random sampling under the same model, budget, and train/test split. Performance is summarised using standard classification or detection metrics, including accuracy, macro- and micro-averaged \(F_1\), mean average precision, and area under the ROC or precision-recall curve; rare-class recall is sometimes reported where class imbalance is severe \citep{kath2024active, zhang2025hybrid, rauch2024towards}. Two common summary values are widely reported: the reduction in labelling effort required to reach a target performance, and the performance gain at a fixed budget. In applied studies, the headline figure is often the percentage reduction in annotation effort \citep{norouzzadeh2021deep, miao2021iterative}. Among the studies in our corpus that report a random sampling baseline, reductions reach 89\% \citep{moller2017active}, with a mean of 64\% (
n=7). Gains reported at a fixed budget are not directly comparable across studies, since they are expressed in whichever metric the study adopts.

This headline figure can mislead when interpreted in isolation. Reduction percentages are inflated on large, redundant datasets: when a multi-million-image camera-trap corpus contains many near-duplicate frames, repeated backgrounds, or otherwise easy samples, a small labelled fraction may suffice even under random sampling. A reduction exceeding 99\% may then reflect the redundancy of the pool and the chosen performance threshold as much as the efficacy of the query strategy \citep{norouzzadeh2021deep}; the absolute reduction conflates the informativeness of the selected samples with the redundancy of the candidate pool. A second weakness is the absence of standardisation: studies report different metrics on different datasets, base models, budgets, and random seeds, which prevents comparison across papers. A third weakness is that a single global metric obscures the quantities most relevant to ecological inference: per-class performance for underrepresented taxa, model calibration, and generalisation under spatiotemporal domain shift \citep{mcewen2025stratified}.

Evaluation is inherently multi-dimensional, and absolute annotation reduction captures only one dimension. \textit{Selection efficiency}, the computational cost of the acquisition step itself, is rarely reported, yet it determines practical applicability: many strategies scale poorly since they require repeated end-to-end retraining, Monte Carlo sampling, or pairwise distance computations over the candidate pool, and per-cycle latency directly affects the usability of human-in-the-loop systems \citep{citovsky2021batch}. The \textit{scale of data} a method can operate over is a second dimension: a strategy validated on a small benchmark may not transfer to the pool sizes of real deployments \citep{citovsky2021batch, beck2023streamline}. \textit{Annotation cost realism} is a third: counting samples ignores the variable cost of annotation, which differs across modalities and between weak and strong labels; we return to this in Section~\ref{sec:active-annotation}.

The random (passive) baseline is the single most important reference point. Its absence from a benchmarking study is a critical flaw because AL does not always outperform random sampling \citep{kholghi2018active, moller2017active, kath2024active, lowell2019practical}: under cold-start conditions, large-batch selection, or distribution shift, query strategies can match or underperform random selection \citep{mittal2019parting}. \citet{kholghi2018active} compared seven query strategies for annotating long-duration environmental recordings and found margin sampling to be the only strategy that never outperformed the baseline; \citet{moller2017active} report uncertainty sampling underperforming random selection over much of the learning curve; and \citet{kath2024active} report a speedup factor of 1.0 for a purely diversity-based strategy, indicating no gain over random selection. In both directions these failures are single-criterion strategies, while the largest gains come from hybrid acquisition functions combining informativeness with diversity, a pattern also seen in the recent BioDCASE challenge \citep{mcewen2026biodcase}. Reporting performance relative to repeated random baselines is the minimum standard for demonstrating that a gain stems from the selection mechanism itself; a single-run baseline is insufficient, since the reported performance of random sampling varies by as much as 13\% between studies using the same dataset and settings \citep{ren2021survey}. A threshold-free summary makes the comparison more portable: either the area under the learning curve expressed as a ratio to the random baseline over a stated budget range, or the speedup factor of \citet{kath2024active}, the fraction of samples an AL strategy requires relative to random sampling, which is invariant to training set size but assumes both strategies converge to the same ceiling performance. Threshold-based reductions should state and motivate the threshold. Publishing the paired learning curves themselves allows any summary statistic to be recomputed. A representative validation or test set must likewise be kept separate from actively selected training samples, because active selection changes the sampled distribution and can bias validation, calibration, and threshold selection \citep{farquhar2021statistical}. In the literature we reviewed, validation data are usually inherited from benchmark splits, and their annotation cost is almost never counted in the reported budget; by the standard of Section~\ref{sec:budget}, few if any studies are budget-complete.

\subsection{Tools and Frameworks}
\label{sec:tools}

Existing software falls into two groups. The first comprises research-oriented AL libraries such as scikit-activeML \citep{herde2025scikit} and comparable toolkits (BaseAL \cite{mcewen2026_baseal}, modAL \cite{danka2018modal}, the DeepAL family \cite{huang2021deepal}): they implement a wide range of query strategies but rely on simulated oracle labelling and lack integration with real annotation workflows, likely because of the complexity of aligning diverse task requirements with different data formats. The second comprises annotation platforms that embed AL or HITL assistance, such as AIDE for camera traps \citep{kellenberger2020aide} and annotation tooling for passive acoustic monitoring \citep{kath2024human, mcewen2024active}. The space between the two groups is the missing loop-closing infrastructure noted in Section~\ref{sec:loop}: no platform has been widely adopted. We argue that what is needed is not more tools but the integration of these methods into the annotation platforms ecologists already use.

The methods surveyed here are applied across different datasets, base models, annotation budgets and modalities, which makes results difficult to compare between studies. Shared tasks address this by fixing the data, the budget and the evaluation protocol so that acquisition strategies are compared under identical conditions. The first \textit{Active Learning for Bioacoustics} task was run in 2026 \citep{mcewen2026biodcase} as part of BioDCASE \citep{stowell2026biodcase}, with participants implementing acquisition functions \citep{magaldi2026determinantal, dubus2026adaptive} over terrestrial and marine mammal datasets within the BaseAL framework \citep{mcewen2026_baseal}.\footnote{The authors organised this task and developed BaseAL.} Version 1.2 of that framework reports annotation and computational cost alongside accuracy for each acquisition method, and supports active testing, so that the labels used for evaluation are selected under a budget rather than assumed to be available. Nothing in the format is specific to bioacoustics, and the same structure could be applied to camera-trap data and other data modalities.

\subsection{Summary}
\label{sec:map-summary}
Taken together, the preceding subsections describe a field configured to demonstrate a single result: that a model of a given quality can be reached with fewer labels than random sampling would require. The datasets are pre-labelled benchmarks, the annotator is a simulated oracle, the dominant strategies optimise classifier improvement, and the dominant metric is the fraction of labels saved. The questions a monitoring programme must answer before acting on model outputs, namely how good the model actually is on its operating distribution and how reliable the ecological estimates built on it are, sit outside this experimental design.

\section{From Label Efficiency to Reliable Inference}
\label{sec:gap}

\subsection{Query-Induced Sampling Bias}
\label{sec:bias}
Query strategies improve label efficiency by oversampling informative regions of the input space, and this is exactly what makes the resulting labelled set unrepresentative \citep{farquhar2021statistical}. Uncertainty sampling concentrates near decision boundaries, so difficult cases are overrepresented: performance estimated on these labels understates the model, and thresholds or calibration maps fitted to them misrepresent the score distributions of the full data stream. The bias is not a side effect of poor implementation; it is the intended behaviour of the selection mechanism.

Part of this bias can be corrected. When samples are selected with known probabilities, importance-weighted estimators recover unbiased risk estimates, and corrected estimators exist for pool-based AL \citep{farquhar2021statistical}. In practice the correction is fragile: many strategies are deterministic top-\(k\) selections without well-defined inclusion probabilities, batch selection couples samples, and weights become extreme when the proposal distribution is far from the population. The dependable alternative is structural: reserve an independently drawn, representative validation set, and treat its size as part of the budget rather than as an afterthought.

A further problem is amplified in monitoring settings: data outlive models. Actively selected datasets are coupled to the model and strategy that produced them, and the gains do not reliably transfer when a different architecture is trained on the same labels \citep{lowell2019practical}. Monitoring archives are reused across model generations, so datasets built through AL should, at a minimum, record the selection mechanism in their metadata so that later users can assess and mitigate bias.

\subsection{Validating Under a Shared Budget}
\label{sec:validation}
What validation requires is a labelled sample set whose distribution matches the conditions under which the model will operate, sized to give acceptable precision for the quantities that matter, and stratified enough to expose per-class and per-site failure. The stopping criterion (Section~\ref{sec:loop}), threshold selection, and every downstream analysis all require validation data. Yet in the reviewed literature, validation data are usually inherited from benchmark splits and their label cost is almost never counted against the reported budget; outside pre-labelled benchmarks, that cost falls squarely on a monitoring programme's own experts.

The validation budget can itself be spent efficiently. Active testing selects which model outputs an expert should verify so that performance estimates reach a target precision with fewer verified labels \citep{kossen2021active}, and sequential variants stop verification once confidence intervals are acceptably narrow \citep{perez2024discount, perez2024human}. Stratified designs allocate verification effort across locations, seasons, or classes \citep{mcewen2025stratified}. A practical recipe follows from this: a two-part budget, \(B_{\mathrm{train}}\) spent on actively selected labels for training and \(B_{\mathrm{val}}\) on representatively drawn (or actively tested) labels for evaluation, with the split between the two reported.

\subsection{Active Annotation}
\label{sec:active-annotation}

The preceding sections treat AL as a question of which samples to label. The cost of each label, however, also depends on how the annotator interacts with the selected sample. It is therefore useful to decompose AL into two components. \textit{Active sampling} is the selection of samples for annotation, the component the rest of this review refers to simply as AL. \textit{Active annotation} concerns how the selected samples are labelled efficiently, for example by presenting the expert with tentative machine labels, rough bounding boxes, or onset and offset times to correct. Both components aim to reduce labelling effort; the AL literature concentrates on the first and treats the cost of each label as fixed.

Active annotation methods include change-point detection and tentative bounding boxes \cite{martinsson2024weak}, extraction and ordering of segments \cite{mcewen2024active, mcewen2023improved}, human-in-the-loop tools for annotating passive acoustic monitoring datasets \cite{kath2024human}, and a broad category of human-interface considerations that improve annotation efficiency. Because the expert budget is measured in time, savings from active annotation add to savings from active sampling, and a budget accounted in expert time captures both. A key challenge for human-in-the-loop methods is the data infrastructure and tooling for closed-loop, iterative labelling. While various tools exist, they have not been widely adopted.

\section{Conclusions}
Machine learning has enabled radical acceleration of various types of monitoring in ecology. However, it relies on trained models and thus retains a bottleneck of expert labelling for training and validating systems. This bottleneck is important as it is often a rate-limiting factor in applied biodiversity monitoring. Active learning, reviewed here, offers the best way to resolve this issue, selecting data in ways which theoretically and empirically make the best use of ecologists' time budgets.

Active learning is most often tested in a traditional way: across acoustic and image applications, fewer labels can suffice to reach a given level of predictive performance. This reduction can be dramatic and represents a key benefit of AL over classical training of recognition algorithms. Our map of 164 studies, however, shows that this evidence comes almost entirely from simulations on pre-labelled benchmarks. The annotator is an oracle, the validation set is free, and the fraction of labels saved is the headline figure. Monitoring programmes face a broader problem that a limited expert budget must simultaneously support model training, model validation, and the ecological estimates built on model outputs, and AL gains its efficiency precisely by distorting the sampling distribution, which disqualifies the labels it buys from serving the latter two purposes.

Some of this can be acted on now. Practitioners can reserve an independently drawn, representative validation set and count it in the budget, report results against repeated random baselines, record the selection mechanism in dataset metadata, and measure costs in expert time rather than sample counts. The research agenda is also clear: query objectives tailored to ecological quantities, formal budget allocation across training and validation, correction methods for query-induced bias in ecological inference, and the nearly empty territories of bats, insects, fish, and multimodal monitoring. The standard by which AL becomes genuinely useful for biodiversity monitoring is not how many labels it saves, but whether, under a fixed expert budget, it delivers a trustworthy model and trustworthy ecological inference.

\section*{Acknowledgement(s)}
BM and DS are funded by Biodiversa+, the European Biodiversity Partnership, in the context of the Towards a Transnational Acoustic Biodiversity MOnitoring Network (TABMON) project under the 2022-2023 BiodivMon joint call. It was co-funded by the European Commission (GA ref. 101052342) and the following funding organisations: Norwegian Research Council (project number 350977), l’Agence Nationale de la Recherche (ANR-23-EBIP-0010), l’Office français de la biodiversité (OFB-23-1865), Dutch Research Council (2023/NWA/01580460), and la Agencia Estatal de Investigación (PCI2024-153427). SZ is funded by the Bioacoustic AI Doctoral Network (European Union Marie Sk{\l}odowska-Curie Action) under grant agreement No 101116715.

\section*{Disclosure statement}

The author(s) declare no competing interests. Generative AI was used to aid in the extraction of information from literature and was used overall to reformulate some sentences and paragraphs. All AI generate content has been manually checked by the authors. Usage of AI is detailed in the manuscript.

\section*{Notes on contributor(s)}
BM conceptualised the review, curated the literature and produced the visualisations. BM and SZ conducted the analysis and wrote the manuscript, contributing equally to the writing. DS reviewed and edited the manuscript.

\bibliographystyle{apacite}
\bibliography{references}

@article{dubus2026adaptive,
  title={Adaptive Diversity-Uncertainty Active Learning with Redundancy Control for Bioacoustic Event Classification},
  author={Dubus, Gabriel and Magaldi, Hugo and Gros-Martial, Anatole},
  journal={arXiv preprint arXiv:2607.04868},
  year={2026}
}

@article{magaldi2026determinantal,
  title={Determinantal point process sampling for bioacoustic active learning},
  author={Magaldi, Hugo and Dubus, Gabriel},
  journal={arXiv preprint arXiv:2607.06063},
  year={2026}
}

@article{mcewen2026biodcase,
  title={BioDCASE: Active Learning for Bioacoustics},
  author={McEwen, Ben and Kurinchi-Vendhan, Rupa and Zhang, Shiqi and Rauch, Lukas and Herde, Marek and Beery, Sara},
  journal={arXiv preprint arXiv:2609.15255},
  year={2026}
}

@article{stowell2026biodcase,
  title={BioDCASE: Using data challenges to make community advances in computational bioacoustics},
  author={Stowell, Dan and Vida{\~n}a-Vila, Ester and Nolasco, Ines and McEwen, Ben and Jean-Labadye, Lucie and Benhamadi, Yasmine and Dubus, Gabriel and Hoffman, Benjamin and Linhart, Pavel and Morandi, Ilaria and others},
  journal={bioRxiv},
  pages={2026--04},
  year={2026},
  publisher={Cold Spring Harbor Laboratory}
}

@article{huang2021deepal,
  title={Deepal: Deep active learning in python},
  author={Huang, Kuan-Hao},
  journal={arXiv preprint arXiv:2111.15258},
  year={2021}
}

@article{danka2018modal,
  title={modAL: A modular active learning framework for Python},
  author={Danka, Tivadar and Horvath, Peter},
  journal={arXiv preprint arXiv:1805.00979},
  year={2018}
}

@software{mcewen2026_baseal,
  author       = {Ben McEwen and
                  Shiqi Zhang},
  title        = {BaseAL: Release v1.2.0},
  month        = aug,
  year         = 2026,
  publisher    = {Zenodo},
  doi          = {10.5281/zenodo.21806641},
  url          = {https://doi.org/10.5281/zenodo.21806641},
}

@inproceedings{gemmeke2017audio,
  title={Audio set: An ontology and human-labeled dataset for audio events},
  author={Gemmeke, Jort F and Ellis, Daniel PW and Freedman, Dylan and Jansen, Aren and Lawrence, Wade and Moore, R Channing and Plakal, Manoj and Ritter, Marvin},
  booktitle={2017 IEEE international conference on acoustics, speech and signal processing (ICASSP)},
  pages={776--780},
  year={2017},
  organization={IEEE}
}

@article{swanson2015snapshot,
  title={Snapshot Serengeti, high-frequency annotated camera trap images of 40 mammalian species in an African savanna},
  author={Swanson, Alexandra and Kosmala, Margaret and Lintott, Chris and Simpson, Robert and Smith, Arfon and Packer, Craig},
  journal={Scientific data},
  volume={2},
  number={1},
  pages={150026},
  year={2015},
  publisher={Nature Publishing Group}
}

@article{canas2023dataset,
  title={A dataset for benchmarking Neotropical anuran calls identification in passive acoustic monitoring},
  author={Ca{\~n}as, Juan Sebasti{\'a}n and Toro-G{\'o}mez, Mar{\'\i}a Paula and Sugai, Larissa Sayuri Moreira and Ben{\'\i}tez Restrepo, Hern{\'a}n Dar{\'\i}o and Rudas, Jorge and Posso Bautista, Breyner and Toledo, Luis Felipe and Dena, Simone and Domingos, Ad{\~a}o Henrique Rosa and de Souza, Franco Leandro and others},
  journal={Scientific Data},
  volume={10},
  number={1},
  pages={771},
  year={2023},
  publisher={Nature Publishing Group UK London}
}

@inproceedings{rauch2025birdset,
  author    = {Rauch, Lukas and Schwinger, Raphael and Wirth, Moritz and Heinrich, Ren{\'e} and Huseljic, Denis and Herde, Marek and Lange, Jonas and Kahl, Stefan and Sick, Bernhard and Tomforde, Sven and Scholz, Christoph},
  title     = {{BirdSet}: A Large-Scale Dataset for Audio Classification in Avian Bioacoustics},
  booktitle = {The Thirteenth International Conference on Learning Representations},
  year      = {2025},
  url       = {https://openreview.net/forum?id=dRXxFEY8ZE}
}

@article{kurinchi2026finding,
  title={Finding Needles in the Haystack: Transductive Active Labeling in Ecology},
  author={Kurinchi-Vendhan, Rupa and Beery, Sara},
  journal={arXiv preprint arXiv:2606.03821},
  year={2026}
}

@techreport{settles2009active,
  title       = {Active learning literature survey},
  author      = {Settles, Burr},
  year        = {2009},
  institution = {University of Wisconsin-Madison Department of Computer Sciences},
  address     = {Madison, WI},
  number      = {1648}
}

@inproceedings{kossen2021active,
  title={Active testing: Sample-efficient model evaluation},
  author={Kossen, Jannik and Farquhar, Sebastian and Gal, Yarin and Rainforth, Tom},
  booktitle={International Conference on Machine Learning},
  pages={5753--5763},
  year={2021},
  organization={PMLR}
}

@inproceedings{alawad2022adaptive,
  title={Adaptive sampling for efficient acoustic noise monitoring: An incremental learning approach},
  author={Alawad, Faiga},
  booktitle={2022 IEEE International Conferences on Internet of Things (iThings) and IEEE Green Computing \& Communications (GreenCom) and IEEE Cyber, Physical \& Social Computing (CPSCom) and IEEE Smart Data (SmartData) and IEEE Congress on Cybermatics (Cybermatics)},
  pages={176--184},
  year={2022},
  organization={IEEE}
}

@article{balantic2019temporally,
  title={Temporally adaptive acoustic sampling to maximize detection across a suite of focal wildlife species},
  author={Balantic, Cathleen and Donovan, Therese},
  journal={Ecology and Evolution},
  volume={9},
  number={18},
  pages={10582--10600},
  year={2019},
  publisher={Wiley Online Library}
}

@inproceedings{lowell2019practical,
  title={Practical obstacles to deploying active learning},
  author={Lowell, David and Lipton, Zachary C and Wallace, Byron C},
  booktitle={Proceedings of the 2019 Conference on Empirical Methods in Natural Language Processing and the 9th International Joint Conference on Natural Language Processing (EMNLP-IJCNLP)},
  pages={21--30},
  year={2019}
}

@article{van2023active,
  title={An active learning framework and assessment of inter-annotator agreement facilitate automated recogniser development for vocalisations of a rare species, the southern black-throated finch (Poephila cincta cincta)},
  author={van Osta, John M and Dreis, Brad and Meyer, Ed and Grogan, Laura F and Castley, J Guy},
  journal={Ecological Informatics},
  volume={77},
  pages={102233},
  year={2023},
  publisher={Elsevier}
}

@inproceedings{arthur2007k,
  title={k-means++: The advantages of careful seeding},
  author={Arthur, David and Vassilvitskii, Sergei},
  booktitle={Soda},
  volume={7},
  pages={1027--1035},
  year={2007}
}

@article{cretois2026tabmon,
  title={TABMON: Design and deployment of a transnational passive acoustic monitoring network for European birds},
  author={Cretois, Benjamin and Rosten, Carolyn M and Wiel, Julia and Barile, Cynthia and McEwen, Ben and Bernard, Corentin and Boom, Michiel P and Bota, Gerard and Brotons, Llu{\`\i}s and Serrano-Davies, Eva and others},
  journal={Methods in Ecology and Evolution},
  publisher={Wiley Online Library},
  year={2026}
}

@inproceedings{farquhar2021statistical,
  title={On statistical bias in active learning: How and when to fix it},
  author={Farquhar, Sebastian and Gal, Yarin and Rainforth, Tom},
  booktitle={International Conference on Learning Representations},
  year={2021}
}

@article{citovsky2021batch,
  title={Batch active learning at scale},
  author={Citovsky, Gui and DeSalvo, Giulia and Gentile, Claudio and Karydas, Lazaros and Rajagopalan, Anand and Rostamizadeh, Afshin and Kumar, Sanjiv},
  journal={Advances in Neural Information Processing Systems},
  volume={34},
  pages={11933--11944},
  year={2021}
}

@article{kholghi2018active,
  title={Active learning for classifying long-duration audio recordings of the environment},
  author={Kholghi, Mahnoosh and Phillips, Yvonne and Towsey, Michael and Sitbon, Laurianne and Roe, Paul},
  journal={Methods in Ecology and Evolution},
  volume={9},
  number={9},
  pages={1948--1958},
  year={2018},
  publisher={Wiley Online Library}
}

@article{lange2023active,
  title={Active learning-based species range estimation},
  author={Lange, Christian and Cole, Elijah and Van Horn, Grant and Mac Aodha, Oisin},
  journal={Advances in neural information processing systems},
  volume={36},
  pages={41892--41913},
  year={2023}
}

@article{miao2021iterative,
  title={Iterative human and automated identification of wildlife images},
  author={Miao, Zhongqi and Liu, Ziwei and Gaynor, Kaitlyn M and Palmer, Meredith S and Yu, Stella X and Getz, Wayne M},
  journal={Nature Machine Intelligence},
  volume={3},
  number={10},
  pages={885--895},
  year={2021},
  publisher={Nature Publishing Group UK London}
}

@article{alksne2026application,
  title={Application of deep learning to estimate blue and fin whale call density in the southern California Current Ecosystem},
  author={Alksne, Michaela N and Roch, Marie A and Frasier, Kaitlin E and Hildebrand, John A and Andres, Shane and Adesanya, Dolapo and Baggett, Lauren M and Jones, Joshua M and Liu, Xiaobai and {\v{S}}irovi{\'c}, Ana and others},
  year={2026},
  journal={preprint}
}

@article{kather2024development,
  title={Development of a machine learning detector for North Atlantic humpback whale song},
  author={Kather, Vincent and Seipel, Fabian and Berges, Benoit and Davis, Genevieve and Gibson, Catherine and Harvey, Matt and Henry, Lea-Anne and Stevenson, Andrew and Risch, Denise},
  journal={The Journal of the Acoustical Society of America},
  volume={155},
  number={3},
  pages={2050--2064},
  year={2024},
  publisher={AIP Publishing}
}

@inproceedings{ayers2021reducing,
  title={Reducing the barriers of acquiring ground-truth from biodiversity rich audio datasets using intelligent sampling techniques},
  author={Ayers, Jacob and Perry, Sean and Tiwari, Vaibhav and Blue, Mugen and Balaji, Nishant and Schurgers, Curt and Kastner, Ryan and Tobler, Mathias and Ingram, Ian},
  booktitle={Tackling Climate Change with AI Workshop, Conference on Neural Information Processing Systems},
  year={2021}
}

@inproceedings{martinsson2024weak,
  title={From weak to strong sound event labels using adaptive change-point detection and active learning},
  author={Martinsson, John and Mogren, Olof and Sandsten, Maria and Virtanen, Tuomas},
  booktitle={2024 32nd European Signal Processing Conference (EUSIPCO)},
  pages={902--906},
  year={2024},
  organization={IEEE}
}

@article{rauch2024towards,
  title={Towards deep active learning in avian bioacoustics},
  author={Rauch, Lukas and Huseljic, Denis and Wirth, Moritz and Decke, Jens and Sick, Bernhard and Scholz, Christoph},
  journal={arXiv preprint arXiv:2406.18621},
  year={2024}
}

@article{kath2024leveraging,
  title={Leveraging transfer learning and active learning for data annotation in passive acoustic monitoring of wildlife},
  author={Kath, Hannes and Serafini, Patricia P and Campos, Ivan B and Gouv{\^e}a, Thiago S and Sonntag, Daniel},
  journal={Ecological Informatics},
  volume={82},
  pages={102710},
  year={2024},
  publisher={Elsevier}
}

@article{mcewen2024active,
  title={Active few-shot learning for rare bioacoustic feature annotation},
  author={McEwen, Ben and Soltero, Kaspar and Gutschmidt, Stefanie and Bainbridge-Smith, Andrew and Atlas, James and Green, Richard},
  journal={Ecological Informatics},
  volume={82},
  pages={102734},
  year={2024},
  publisher={Elsevier}
}

@article{lindholm2025aggregation,
  title={Aggregation Strategies for Efficient Annotation of Bioacoustic Sound Events Using Active Learning},
  author={Lindholm, Richard and Marklund, Oscar and Mogren, Olof and Martinsson, John},
  journal={arXiv preprint arXiv:2503.02422},
  year={2025}
}

@article{mcewen2025stratified,
  title={Stratified Active Learning for Spatiotemporal Generalisation in Bioacoustic Monitoring},
  author={McEwen, Ben and Bernard, Corentin and Stowell, Dan},
  journal={bioRxiv},
  pages={2025--09},
  year={2025},
  publisher={Cold Spring Harbor Laboratory}
}

@inproceedings{kath2024human,
  title={A human-in-the-loop tool for annotating passive acoustic monitoring datasets},
  author={Kath, Hannes and Gouv{\^e}a, Thiago S and Sonntag, Daniel},
  booktitle={German Conference on Artificial Intelligence (K{\"u}nstliche Intelligenz)},
  pages={341--345},
  year={2024},
  organization={Springer}
}

@article{bordoux2026rapid,
  title={Rapid fish sound detection using human-in-the-loop deep learning},
  author={Bordoux, Valentin and Parcerisas, Clea and Martin, J{\"a}lmby and Elisabeth, Debusschere and Murk, Albertinka J and van der Ven, Rosa M},
  journal={Ecological Informatics},
  pages={103874},
  year={2026},
  publisher={Elsevier}
}

@inproceedings{kath2024active,
  title={Active learning in multi-label classification of bioacoustic data},
  author={Kath, Hannes and Gouv{\^e}a, Thiago S and Sonntag, Daniel},
  booktitle={German Conference on Artificial Intelligence (K{\"u}nstliche Intelligenz)},
  pages={114--127},
  year={2024},
  organization={Springer}
}

@inproceedings{kath2025speeding,
  title={Speeding Up Bioacoustic Data Analysis: Fine-Tuning Deep Models with Active Learning for Efficient Wildlife Detection},
  author={Kath, Hannes and Gouv{\^e}a, Thiago S and Sonntag, Daniel},
  booktitle={Proceedings of the 2025 International Conference on Information Technology for Social Good (GoodIT)},
  pages={200--203},
  year={2025},
  publisher={ACM}
}

@inproceedings{mcewen2023improved,
  title={An improved computational bioacoustic monitoring approach for sparse features detection},
  author={McEwen, Ben and Soltero, Kaspar and Gutschmidt, Stefanie and Bainbridge-Smith, Andrew and Atlas, James and Green, Richard},
  booktitle={Proceedings of Meetings on Acoustics},
  volume={52},
  number={1},
  pages={022002},
  year={2023},
  organization={Acoustical Society of America}
}

@article{dumoulin2025search,
  title={The Search for Squawk: Agile Modeling in Bioacoustics},
  author={Dumoulin, Vincent and Stretcu, Otilia and Hamer, Jenny and Harrell, Lauren and Laber, Rob and Larochelle, Hugo and van Merri{\"e}nboer, Bart and Navine, Amanda and Hart, Patrick and Williams, Ben and others},
  journal={arXiv preprint arXiv:2505.03071},
  year={2025}
}

@inproceedings{kath2024active_socialgood,
  title={Active and Transfer Learning for Efficient Identification of Species in Multi-Label Bioacoustic Datasets},
  author={Kath, Hannes and Gouv{\^e}a, Thiago S and Sonntag, Daniel},
  booktitle={Proceedings of the 2024 International Conference on Information Technology for Social Good},
  pages={22--25},
  year={2024}
}

@article{zhang2025hybrid,
  title={Hybrid Disagreement-Diversity Active Learning for Bioacoustic Sound Event Detection},
  author={Zhang, Shiqi and Virtanen, Tuomas},
  journal={arXiv preprint arXiv:2505.20956},
  year={2025}
}

@article{kitzes2025integrating,
  title={Integrating {AI} models into ecological research workflows: The case of terrestrial bioacoustics},
  author={Kitzes, Justin and Chronister, Lauren and Czarnecki, Chapin and Fiss, Cameron and Freeland-Haynes, Louis and Goodman, Brooke D and Lapp, Sam and Lyon, R Patrick and Nossan, Hannah and Rhinehart, Tessa A and Ruiz Guzman, Santiago and Syunkova, Alexandra and Viotti, Leonardo},
  journal={Methods in Ecology and Evolution},
  volume={17},
  number={2},
  pages={257--271},
  year={2026},
  doi={10.1111/2041-210X.70133},
  publisher={Wiley Online Library}
}

@article{stowell2022computational,
  title={Computational bioacoustics with deep learning: a review and roadmap},
  author={Stowell, Dan},
  journal={PeerJ},
  volume={10},
  pages={e13152},
  year={2022},
  publisher={PeerJ Inc.}
}

@article{schwinger2025foundation,
  title={Foundation Models for Bioacoustics--a Comparative Review},
  author={Schwinger, Raphael and Zadeh, Paria Vali and Rauch, Lukas and Kurz, Mats and Hauschild, Tom and Lapp, Sam and Tomforde, Sven},
  journal={arXiv preprint arXiv:2508.01277},
  year={2025}
}

@article{qian2017active,
  title={Active learning for bird sound classification via a kernel-based extreme learning machine},
  author={Qian, Kun and Zhang, Zixing and Baird, Alice and Schuller, Bj{\"o}rn},
  journal={The Journal of the Acoustical Society of America},
  volume={142},
  number={4},
  pages={1796--1804},
  year={2017},
  publisher={AIP Publishing}
}

@article{mohaimenuzzaman2023deep,
  title={Deep active audio feature learning in resource-constrained environments},
  author={Mohaimenuzzaman, Md and Bergmeir, Christoph and Meyer, Bernd},
  journal={arXiv preprint arXiv:2308.13201},
  year={2023}
}

@article{williams2025using,
  title={Using tropical reef, bird and unrelated sounds for superior transfer learning in marine bioacoustics},
  author={Williams, Ben and van Merri{\"e}nboer, Bart and Dumoulin, Vincent and Hamer, Jenny and Fleishman, Abram B and McKown, Matthew and Munger, Jill and Rice, Aaron N and Lillis, Ashlee and White, Clemency and others},
  journal={Philosophical Transactions B},
  volume={380},
  number={1928},
  pages={20240280},
  year={2025},
  publisher={The Royal Society}
}

@article{rauch2023active,
  title={Active bird2vec: towards end-to-end bird sound monitoring with transformers},
  author={Rauch, Lukas and Schwinger, Raphael and Wirth, Moritz and Sick, Bernhard and Tomforde, Sven and Scholz, Christoph},
  journal={arXiv preprint arXiv:2308.07121},
  year={2023}
}

@article{ghani2023global,
  title={Global birdsong embeddings enable superior transfer learning for bioacoustic classification},
  author={Ghani, Burooj and Denton, Tom and Kahl, Stefan and Klinck, Holger},
  journal={Scientific Reports},
  volume={13},
  number={1},
  pages={22876},
  year={2023},
  publisher={Nature Publishing Group UK London}
}

@article{bernard2025data,
  title={Data-driven Sampling Strategies for Fine-Tuning Bird Detection Models},
  author={Bernard, Corentin and McEwen, Ben and Cretois, Benjamin and Glotin, Herv{\'e} and Stowell, Dan and Marxer, Ricard},
  journal={bioRxiv},
  pages={2025--10},
  year={2025},
  publisher={Cold Spring Harbor Laboratory}
}

@article{rauch2025can,
  title={Can Masked Autoencoders Also Listen to Birds?},
  author={Rauch, Lukas and Heinrich, Ren{\'e} and Moummad, Ilyass and Joly, Alexis and Sick, Bernhard and Scholz, Christoph},
  journal={arXiv preprint arXiv:2504.12880},
  year={2025}
}

@article{allen2021convolutional,
  title={A convolutional neural network for automated detection of humpback whale song in a diverse, long-term passive acoustic dataset},
  author={Allen, Ann N and Harvey, Matt and Harrell, Lauren and Jansen, Aren and Merkens, Karlina P and Wall, Carrie C and Cattiau, Julie and Oleson, Erin M},
  journal={Frontiers in Marine Science},
  volume={8},
  pages={607321},
  year={2021},
  publisher={Frontiers Media SA}
}

@inproceedings{bellafkir2023edge,
  title={Edge-Based Bird Species Recognition via Active Learning},
  author={Bellafkir, Hicham and Vogelbacher, Markus and Schneider, Daniel and M{\"u}hling, Markus and Korfhage, Nikolaus and Freisleben, Bernd},
  booktitle={International Conference on Networked Systems},
  pages={17--34},
  year={2023},
  organization={Springer}
}

@article{ghani2025impact,
  title={Impact of transfer learning methods and dataset characteristics on generalization in birdsong classification},
  author={Ghani, Burooj and Kalkman, Vincent J and Planqu{\'e}, Bob and Vellinga, Willem-Pier and Gill, Lisa and Stowell, Dan},
  journal={Scientific Reports},
  volume={15},
  number={1},
  pages={16273},
  year={2025},
  publisher={Nature Publishing Group UK London}
}

@inproceedings{burnsperch,
  title={Perch 2.0 transfers '{whale}' to underwater tasks},
  author={Burns, Andrea and Harrell, Lauren and van Merri{\"e}nboer, Bart and Dumoulin, Vincent and Hamer, Jenny and Denton, Tom},
  booktitle={The Thirty-Ninth Annual Conference on Neural Information Processing Systems workshop: AI for non-human animal communication},
  year={2025}
}

@article{van2024seasonal,
  title={Seasonal habitat use of the southern black-throated finch},
  author={van Osta, JM and Dreis, B and Grogan, LF and Castley, JG and van Osta, John},
  journal={Statement of originality},
  pages={148},
  year={2024},
  publisher={Griffith University}
}

@article{williams2025evidence,
  title={Evidence of ecosystem process recovery across a large-scale coral reef restoration programme using AI accelerated soundscape analysis},
  author={Williams, Ben and Mars Coral Restoration Project Monitoring Team and Naseem, Aya and Nava, Gaby and Roberts, Angus and Nicholson, Freda and du Luart, Aimee and Erasmus, Dave and Stoole, Ollie and Whittick, Alistair and others},
  journal={bioRxiv},
  pages={2025--09},
  year={2025},
  publisher={Cold Spring Harbor Laboratory}
}

@article{kozlova2025dlc2action,
  title={DLC2Action: A Deep Learning-based Toolbox for Automated Behavior Segmentation},
  author={Kozlova, Elizaveta and Bonnetto, Andy and Mathis, Alexander},
  journal={bioRxiv},
  pages={2025--09},
  year={2025},
  publisher={Cold Spring Harbor Laboratory}
}

@article{erbs2023towards,
  title={Towards automated long-term acoustic monitoring of endangered river dolphins: a case study in the Brazilian Amazon floodplains},
  author={Erbs, Florence and Gaona, Marina and van der Schaar, Mike and Zaugg, Serge and Ramalho, Emiliano and Houser, Dorian and Andr{\'e}, Michel},
  journal={Scientific reports},
  volume={13},
  number={1},
  pages={10801},
  year={2023},
  publisher={Nature Publishing Group UK London}
}

@article{van2022emerging,
  title={Emerging technologies revolutionise insect ecology and monitoring},
  author={Van Klink, Roel and August, Tom and Bas, Yves and Bodesheim, Paul and Bonn, Aletta and Foss{\o}y, Frode and H{\o}ye, Toke T and Jongejans, Eelke and Menz, Myles HM and Miraldo, Andreia and others},
  journal={Trends in ecology \& evolution},
  volume={37},
  number={10},
  pages={872--885},
  year={2022},
  publisher={Elsevier}
}

@article{martinsson2025accuracy,
  title={The Accuracy Cost of Weakness: A Theoretical Analysis of Fixed-Segment Weak Labeling for Events in Time},
  author={Martinsson, John and Virtanen, Tuomas and Sandsten, Maria and Mogren, Olof},
  journal={arXiv preprint arXiv:2502.09363},
  year={2025}
}

@article{allen2024bryde,
  title={Bryde’s whales produce Biotwang calls, which occur seasonally in long-term acoustic recordings from the central and western North Pacific},
  author={Allen, Ann N and Harvey, Matt and Harrell, Lauren and Wood, Megan and Szesciorka, Angela R and McCullough, Jennifer LK and Oleson, Erin M},
  journal={Frontiers in Marine Science},
  volume={11},
  pages={1394695},
  year={2024},
  publisher={Frontiers Media SA}
}

@article{sheard2024emerging,
  title={Emerging technologies in citizen science and potential for insect monitoring},
  author={Sheard, Julie Koch and Adriaens, Tim and Bowler, Diana E and B{\"u}ermann, Andrea and Callaghan, Corey T and Camprasse, Elodie CM and Chowdhury, Shawan and Engel, Thore and Finch, Elizabeth A and Von G{\"o}nner, Julia and others},
  journal={Philosophical Transactions of the Royal Society B},
  volume={379},
  number={1904},
  pages={20230106},
  year={2024},
  publisher={The Royal Society}
}

@inproceedings{chowdhury2024active,
  title={Active learning strategy using contrastive learning and k-means for aquatic invasive species recognition},
  author={Chowdhury, Shaif and Hamerly, Greg and McGarrity, Monica},
  booktitle={Proceedings of the IEEE/CVF Winter Conference on Applications of Computer Vision},
  pages={848--858},
  year={2024}
}

@article{norouzzadeh2021deep,
  title={A deep active learning system for species identification and counting in camera trap images},
  author={Norouzzadeh, Mohammad Sadegh and Morris, Dan and Beery, Sara and Joshi, Neel and Jojic, Nebojsa and Clune, Jeff},
  journal={Methods in ecology and evolution},
  volume={12},
  number={1},
  pages={150--161},
  year={2021},
  publisher={Wiley Online Library}
}

@article{bothmann2023automated,
  title={Automated wildlife image classification: An active learning tool for ecological applications},
  author={Bothmann, Ludwig and Wimmer, Lisa and Charrakh, Omid and Weber, Tobias and Edelhoff, Hendrik and Peters, Wibke and Nguyen, Hien and Benjamin, Caryl and Menzel, Annette},
  journal={Ecological Informatics},
  volume={77},
  pages={102231},
  year={2023},
  publisher={Elsevier}
}

@book{monarch2021human,
  title={Human-in-the-Loop Machine Learning: Active learning and annotation for human-centered AI},
  author={Monarch, Robert Munro},
  year={2021},
  publisher={Simon and Schuster}
}

@article{bodesheim2022pre,
  title={Pre-trained models are not enough: active and lifelong learning is important for long-term visual monitoring of mammals in biodiversity research—individual identification and attribute prediction with image features from deep neural networks and decoupled decision models applied to elephants and great apes},
  author={Bodesheim, Paul and Blunk, Jan and Koerschens, Matthias and Brust, Clemens-Alexander and Kaeding, Christoph and Denzler, Joachim},
  journal={Mammalian Biology},
  volume={102},
  number={3},
  pages={875--897},
  year={2022},
  publisher={Springer}
}

@article{mononen2025active,
  title={An active ensemble classifier for detecting animal sequences from global camera trap data},
  author={Mononen, Tommi and Hardwick, Bess and Alcobia, Sandra and Barrett, Adrian and Blagoev, Gergin A and Boyer, St{\'e}phane and Gon{\c{c}}alves, Paula and Gottsberger, Brigitte and Groner, Elli and Ho, Chris CY and others},
  journal={Methods in Ecology and Evolution},
  volume={16},
  number={10},
  pages={2500--2516},
  year={2025},
  publisher={Wiley Online Library}
}

@inproceedings{auer2021minimizing,
  title={Minimizing the annotation effort for detecting wildlife in camera trap images with active learning},
  author={Auer, Daphne and Bodesheim, Paul and Fiderer, Christian and Heurich, Marco and Denzler, Joachim},
  booktitle={INFORMATIK 2021},
  pages={547--564},
  year={2021},
  organization={Gesellschaft f{\"u}r Informatik, Bonn}
}

@article{sani2025active,
  title={Active Learning for Animal Re-Identification with Ambiguity-Aware Sampling},
  author={Sani, Depanshu and Khurana, Mehar and Anand, Saket},
  journal={arXiv preprint arXiv:2511.06658},
  year={2025}
}

@article{nguyen2025model,
  title={A model-agnostic active learning approach for animal detection from camera traps},
  author={Nguyen, Thi Thu Thuy and Nguyen, Duc Thanh},
  journal={arXiv preprint arXiv:2507.06537},
  year={2025}
}

@article{fabian2023multimodal,
  title={Multimodal foundation models for zero-shot animal species recognition in camera trap images},
  author={Fabian, Zalan and Miao, Zhongqi and Li, Chunyuan and Zhang, Yuanhan and Liu, Ziwei and Hern{\'a}ndez, Andr{\'e}s and Montes-Rojas, Andr{\'e}s and Escucha, Rafael and Siabatto, Laura and Link, Andr{\'e}s and others},
  journal={arXiv preprint arXiv:2311.01064},
  year={2023}
}

@inproceedings{kulits2021elephantbook,
  title={ElephantBook: A semi-automated human-in-the-loop system for elephant re-identification},
  author={Kulits, Peter and Wall, Jake and Bedetti, Anka and Henley, Michelle and Beery, Sara},
  booktitle={Proceedings of the 4th ACM SIGCAS Conference on Computing and Sustainable Societies},
  pages={88--98},
  year={2021}
}

@article{brust2020active,
  title={Active and incremental learning with weak supervision},
  author={Brust, Clemens-Alexander and K{\"a}ding, Christoph and Denzler, Joachim},
  journal={KI-K{\"u}nstliche Intelligenz},
  volume={34},
  number={2},
  pages={165--180},
  year={2020},
  publisher={Springer}
}

@article{beck2023streamline,
  title={Streamline: Streaming active learning for realistic multi-distributional settings},
  author={Beck, Nathan and Kothawade, Suraj and Shenoy, Pradeep and Iyer, Rishabh},
  journal={arXiv preprint arXiv:2305.10643},
  year={2023}
}

@article{tamkin2022active,
  title={Active learning helps pretrained models learn the intended task},
  author={Tamkin, Alex and Nguyen, Dat and Deshpande, Salil and Mu, Jesse and Goodman, Noah},
  journal={Advances in Neural Information Processing Systems},
  volume={35},
  pages={28140--28153},
  year={2022}
}

@article{kellenberger2020aide,
  title={AIDE: Accelerating image-based ecological surveys with interactive machine learning},
  author={Kellenberger, Benjamin and Tuia, Devis and Morris, Dan},
  journal={Methods in Ecology and Evolution},
  volume={11},
  number={12},
  pages={1716--1727},
  year={2020},
  publisher={Wiley Online Library}
}

@inproceedings{perez2024human,
  title={Human-in-the-loop visual re-id for population size estimation},
  author={Perez, Gustavo and Sheldon, Daniel and Van Horn, Grant and Maji, Subhransu},
  booktitle={European Conference on Computer Vision},
  pages={185--202},
  year={2024},
  organization={Springer}
}

@article{ren2021survey,
  title={A survey of deep active learning},
  author={Ren, Pengzhen and Xiao, Yun and Chang, Xiaojun and Huang, Po-Yao and Li, Zhihui and Gupta, Brij B and Chen, Xiaojiang and Wang, Xin},
  journal={ACM computing surveys (CSUR)},
  volume={54},
  number={9},
  pages={1--40},
  year={2021},
  publisher={ACM New York, NY}
}

@misc{zhang2025evaluating,
  title={Evaluating Active Learning and Classifiers on Laying Hens’ Motion Data of 27-Behavioral Classes},
  author={Zhang, Guihao and Fujinami, Kaori and Shimmura, Tsuyoshi},
  year={2025},
  howpublished={Preprints.org}
}

@inproceedings{moller2017active,
  title={Active learning for the classification of species in underwater images from a fixed observatory},
  author={Moller, Torben and Nilssen, Ingunn and Nattkemper, Tim W},
  booktitle={Proceedings of the IEEE International Conference on Computer Vision Workshops},
  pages={2891--2897},
  year={2017}
}

@inproceedings{perez2024discount,
  title={DISCount: counting in large image collections with detector-based importance sampling},
  author={Perez, Gustavo and Maji, Subhransu and Sheldon, Daniel},
  booktitle={Proceedings of the AAAI Conference on Artificial Intelligence},
  volume={38},
  number={20},
  pages={22294--22302},
  year={2024}
}

@incollection{van2025arise,
  title={ARISE: a Dutch dataspace connecting nature and people},
  author={van Ommen Kloeke, Elaine and Kissling, W Daniel and Evans, Julian and Huijbers, Chantal and Kamminga, Jacob and Schouten, Gerard},
  booktitle={Moral design and green technology},
  pages={233--251},
  year={2025},
  publisher={Wageningen Academic}
}

@inproceedings{boinski2020collaborative,
  title={Collaborative Data Acquisition and Learning Support},
  author={Boi{\'n}ski, Tomasz and Szyma{\'n}ski, Julian},
  booktitle={International Conference on Computer Information Systems and Industrial Management},
  pages={220--229},
  year={2020},
  organization={Springer}
}

@inproceedings{kang2021exploiting,
  title={Exploiting proximity search and easy examples to select rare events},
  author={Kang, Daniel and Derhacobian, Alex and Tsuji, Kaoru and Hebert, Trevor and Bailis, Peter and Fukami, Tadashi and Hashimoto, Tatsunori and Sun, Yi and Zaharia, Matei},
  booktitle={NeurIPS Data-Centric AI Workshop 2021},
  year={2021}
}

@article{wen2024feature,
  title={Feature Alignment: Rethinking Efficient Active Learning via Proxy in the Context of Pre-trained Models},
  author={Wen, Ziting and Pizarro, Oscar and Williams, Stefan},
  journal={arXiv preprint arXiv:2403.01101},
  year={2024}
}

@inproceedings{stevens2024bioclip,
  title={Bioclip: A vision foundation model for the tree of life},
  author={Stevens, Samuel and Wu, Jiaman and Thompson, Matthew J and Campolongo, Elizabeth G and Song, Chan Hee and Carlyn, David Edward and Dong, Li and Dahdul, Wasila M and Stewart, Charles and Berger-Wolf, Tanya and others},
  booktitle={Proceedings of the IEEE/CVF conference on computer vision and pattern recognition},
  pages={19412--19424},
  year={2024}
}

@article{herde2025scikit,
  title={scikit-activeml: A Comprehensive and User-friendly Active Learning Library},
  author={Herde, Marek and Pham, Minh Tuan and Kottke, Daniel and Benz, Alexander and L{\"u}hrs, Lukas and Mergard, Pascal and Sandrock, Christoph and Cheng, Jiaying and Roghman, Atal and M{\"u}jde, Mehmet and others},
  year={2025},
  journal={pre-print}
}

@article{kahl2021birdnet,
  title={{BirdNET}: A deep learning solution for avian diversity monitoring},
  author={Kahl, Stefan and Wood, Connor M and Eibl, Maximilian and Klinck, Holger},
  journal={Ecological Informatics},
  volume={61},
  pages={101236},
  year={2021},
  publisher={Elsevier}
}

@inproceedings{hagiwara2023aves,
  title={{AVES}: Animal vocalization encoder based on self-supervision},
  author={Hagiwara, Masato},
  booktitle={ICASSP 2023 -- 2023 IEEE International Conference on Acoustics, Speech and Signal Processing (ICASSP)},
  pages={1--5},
  year={2023},
  organization={IEEE}
}

@article{schafer2024animal2vec,
  title={animal2vec and {MeerKAT}: A self-supervised transformer for rare-event raw audio input and a large-scale reference dataset for bioacoustics},
  author={Sch{\"a}fer-Zimmermann, Julian C and Demartsev, Vlad and Averly, Baptiste and Dhanjal-Adams, Kiran and Duteil, Mathieu and Gall, Gabriella and Fai{\ss}, Marius and Johnson-Ulrich, Lily and Stowell, Dan and Manser, Marta B and others},
  journal={arXiv preprint arXiv:2406.01253},
  year={2024}
}

@inproceedings{chen2022beats,
  title={{BEATs}: Audio pre-training with acoustic tokenizers},
  author={Chen, Sanyuan and Wu, Yu and Wang, Chengyi and Liu, Shujie and Tompkins, Daniel and Chen, Zhuo and Che, Wanxiang and Yu, Xiangzhan and Wei, Furu},
  booktitle={Proceedings of the 40th International Conference on Machine Learning (ICML)},
  volume={202},
  pages={5178--5193},
  year={2023},
  organization={PMLR}
}

@article{beery2019efficient,
  title={Efficient pipeline for camera trap image review},
  author={Beery, Sara and Morris, Dan and Yang, Siyu},
  journal={arXiv preprint arXiv:1907.06772},
  year={2019}
}

@article{hernandez2024pytorchwildlife,
  title={Pytorch-{W}ildlife: A collaborative deep learning framework for conservation},
  author={Hernandez, Andres and Miao, Zhongqi and Vargas, Luisa and Beery, Sara and Dodhia, Rahul and Arbelaez, Pablo and Lavista Ferres, Juan M},
  journal={arXiv preprint arXiv:2405.12930},
  year={2024}
}

@inproceedings{xie2023active,
  title={Active finetuning: Exploiting annotation budget in the pretraining-finetuning paradigm},
  author={Xie, Yichen and Lu, Han and Yan, Junchi and Yang, Xiaokang and Tomizuka, Masayoshi and Zhan, Wei},
  booktitle={Proceedings of the IEEE/CVF Conference on Computer Vision and Pattern Recognition (CVPR)},
  pages={23715--23724},
  year={2023}
}

@inproceedings{hacohen2022typiclust,
  title={Active learning on a budget: Opposite strategies suit high and low budgets},
  author={Hacohen, Guy and Dekel, Avihu and Weinshall, Daphna},
  booktitle={Proceedings of the 39th International Conference on Machine Learning (ICML)},
  volume={162},
  pages={8175--8195},
  year={2022},
  organization={PMLR}
}

@inproceedings{sener2018active,
  title={Active learning for convolutional neural networks: A core-set approach},
  author={Sener, Ozan and Savarese, Silvio},
  booktitle={International Conference on Learning Representations (ICLR)},
  year={2018}
}

@article{mittal2019parting,
  title={Parting with illusions about deep active learning},
  author={Mittal, Sudhanshu and Tatarchenko, Maxim and {\c{C}}i{\c{c}}ek, {\"O}zg{\"u}n and Brox, Thomas},
  journal={arXiv preprint arXiv:1912.05361},
  year={2019}
}

\appendix

\section{Schema}
\label{app:schema}

Each paper was processed and extracted against the 24-field schema below. Fields use fixed response options where possible to allow aggregation across the corpus. Papers judged not relevant (Field 2) were not processed further; a short note recording the reason for exclusion was kept instead. Any field that could not be determined confidently from the text was flagged for manual review by the authors rather than inferred.
 
\definecolor{schemagrey}{gray}{0.92}
\newcommand{\schemagroup}[1]{%
  \rowcolor{schemagrey}\multicolumn{3}{@{\hspace{4pt}}l}{\rule{0pt}{2.6ex}\textbf{#1}} \\[2pt]}
 
{\small
\setlength{\LTcapwidth}{\textwidth}
\renewcommand{\arraystretch}{1.25}
\setlength{\tabcolsep}{5pt}
\begin{xltabular}{\textwidth}{@{}>{\raggedleft\arraybackslash}p{1.4em} >{\RaggedRight\arraybackslash}p{0.27\textwidth} >{\RaggedRight\arraybackslash}X@{}}
\caption{Extraction schema applied to each paper. Options separated by | are mutually exclusive categories.}
\label{tab:schema} \\
\toprule
\textbf{\#} & \textbf{Field} & \textbf{Options and extraction notes} \\
\midrule
\endfirsthead
\multicolumn{3}{@{}l}{\small\textit{Table \thetable{} continued}} \\
\toprule
\textbf{\#} & \textbf{Field} & \textbf{Options and extraction notes} \\
\midrule
\endhead
\midrule
\multicolumn{3}{r@{}}{\small\textit{Continued on next page}} \\
\endfoot
\bottomrule
\endlastfoot
 
\schemagroup{Paper metadata}
1  & Pre-print & Yes | No \\
2  & Relevant & Yes | No [If No, then extraction does not proceed] \\
\midrule
 
\schemagroup{Active learning configuration}
3  & AL type & Pure AL | HITL with AL | HITL without AL | Neither. \newline
  \textit{Pure AL:} model selects samples to label. \newline
  \textit{HITL without AL:} human verification or correction, no active selection. \newline
  \textit{Neither:} passive learning with human annotation. \\
4  & AL scenario & Pool-based | Stream-based | Membership query synthesis | Other \\
5  & Query strategy & Primary selection method: uncertainty (e.g.\ entropy, margin, BALD), diversity (e.g.\ clustering, core-set) or hybrid. \\
6  & Diversification applied & Yes | No. Any diversity-promoting mechanism, as the primary strategy or in combination. \\
7  & Batch vs.\ sequential & Single instance | Batch selection \\
8  & Stopping criterion & Method used, or not reported. \\
9  & Labelling budget & Total annotations used, if reported. \\
10 & Multi-label & Yes | No \\
\midrule
 
\schemagroup{Performance and evaluation}
11 & Improvement over baseline & Yes | No \\
12 & Evaluation metrics & e.g.\ F1, mAP, AUC, accuracy \\
13 & Data reduction & Reduction in labelled samples (e.g.\ \% of pool), or not reported. \\
\midrule
 
\schemagroup{Application domain}
14 & Modality & Acoustic | Image | eDNA | Earth Observation | NLP | Combined \\
15 & Domain & Terrestrial | Marine | Both  | N/A \\
16 & Taxonomic group & e.g.\ birds, mammals, insects \\
\midrule
 
\schemagroup{Technical details}
17 & Transfer learning & Yes | No \\
18 & Base model(s) & Architecture names \\
19 & Dataset(s) & Names and sources \\
20 & Contribution type & Technical | Applied | Both. \newline
  \textit{Technical:} novel sampling methodology. \newline
  \textit{Applied:} existing sampling methods applied to a new problem. \\
\midrule
 
\schemagroup{Context}
21 & Reported motivation & Stated reason for using AL, e.g.\ annotation cost, rare species. \\
22 & Annotation tool & Tool used, created or published, or None. \\
\midrule
 
\schemagroup{Summary}
23 & Brief summary & One to two sentences. \\
24 & Limitations and gaps & Limitations and research gaps identified by the authors or during review. \\
\addlinespace[2pt]
\end{xltabular}
 
\paragraph*{Optional fields.} Where relevant, four further fields were recorded: class imbalance handling, cold-start strategy, treatment of spatial or temporal structure, and whether domain shift across sites or time periods was addressed.

\end{document}